\documentclass{article}

\usepackage{microtype}
\usepackage{graphicx}
\usepackage{subfigure}
\usepackage{booktabs} 

\usepackage{hyperref}

\usepackage[accepted]{icml2021}

\usepackage{amsmath,amsfonts,bm, amsthm}

\def\1{\bm{1}}

\def\vd{{\bm{d}}}

\def\vw{{\bm{w}}}

\DeclareMathAlphabet{\mathsfit}{\encodingdefault}{\sfdefault}{m}{sl}
\SetMathAlphabet{\mathsfit}{bold}{\encodingdefault}{\sfdefault}{bx}{n}

\newcommand{\R}{\mathbb{R}}

\icmltitlerunning{Dense for the Price of Sparse}

\usepackage{tabularx,ragged2e,booktabs}
\newcolumntype{A}{>{\hsize=1.7cm}X}
\newcolumntype{B}{>{\Centering\arraybackslash\hsize=1.1cm}X}
\newcolumntype{C}{>{\Centering\arraybackslash\hsize=1.1cm}X}
\newcolumntype{D}{>{\Centering\arraybackslash\hsize=1.1cm}X}
\newcolumntype{E}{>{\Centering\arraybackslash\hsize=1.2cm}X}
\newcolumntype{F}{>{\Centering\arraybackslash\hsize=1.7cm}X}
\newcolumntype{G}{>{\Centering\arraybackslash\hsize=1.6cm}X}
\newcolumntype{H}{>{\Centering\arraybackslash\hsize=1.2cm}X}
\newcolumntype{I}{>{\Centering\arraybackslash\hsize=1.5cm}X}
\newcolumntype{J}{>{\Centering\arraybackslash\hsize=1.1cm}X}

\usepackage{dsfont}

\begin{document}

\twocolumn[
\icmltitle{Dense for the Price of Sparse: Improved Performance of Sparsely Initialized Networks via a Subspace Offset}





\begin{icmlauthorlist}
\icmlauthor{Ilan Price}{ox,ati}
\icmlauthor{Jared Tanner}{ox,ati}
\end{icmlauthorlist}

\icmlaffiliation{ox}{Mathematical Institute, University of Oxford, Oxford, UK}
\icmlaffiliation{ati}{The Alan Turing Institute, London, UK}

\icmlcorrespondingauthor{Ilan Price}{ilan.price@maths.ox.ac.uk}

\icmlkeywords{}

\vskip 0.3in
]



\printAffiliationsAndNotice{}  

\begin{abstract}
That neural networks may be pruned to high sparsities and retain high accuracy is well established. Recent research efforts focus on pruning immediately after initialization so as to allow the computational savings afforded by sparsity to extend to the training process. In this work, we introduce a new `DCT plus Sparse' layer architecture, which maintains information propagation and trainability even with as little as 0.01\% trainable kernel parameters remaining. We show that standard training of networks built with these layers, and pruned at initialization, achieves state-of-the-art accuracy for extreme sparsities on a variety of benchmark network architectures and datasets. Moreover, these results are achieved using only simple heuristics to determine the locations of the trainable parameters in the network, and thus without having to initially store or compute with the full, unpruned network, as is required by competing prune-at-initialization algorithms. Switching from standard sparse layers to DCT plus Sparse layers does not increase the storage footprint of a network and incurs only a small additional computational overhead.
\end{abstract}

\section{Introduction}
It is well established that neural networks can be pruned extensively while retaining high accuracy; see \cite{blalock2020state, liu2020pruning} for recent reviews. Sparse networks have significant potential benefits in terms of the memory and computational costs of training and applying large networks, as well as the cost of communication between servers and edge devices in the context of federated learning. Consequently research on pruning techniques has garnered significant momentum over the last few years. 

\begin{table*}[h!]
\caption{Summary of state-of-the-art PaI and DST methods in comparison with the DCT plus Sparse (DCTpS) approach presented in this paper. Uniform random pruning is included as baseline, and Iterative Magnitude Pruning (IMP), though not a PaI algorithm, is included for comparison. The table considers only the pruning of network weights, not bias or batchnorm parameters, which is the focus of prior PaI work and of this paper. Let $N$ denote the total number of parameters in the full networks' weights, $P\in (0,1)$ is the global overall density of the weights tensors, and $k$ is the number of iterations used in a PaI algorithm. The computational cost in the table refers to the average cost of each individual matrix-vector product involved in feedforward and convolutional layers, with flattened weights tensor of size $m \times n$ with density $p$, with $q = \max(m,n)$ (see Section \ref{sec: complexity} for more details). If this cost differs between the forward and backward pass, the larger of the two is included. The extra factor of $\frac{1}{\Delta T}mn$ for RigL comes from the necessity to compute full gradients every $\Delta T$ steps.   We report the drop in accuracy for the case of ResNet50 applied to CIFAR100 at different global sparsities, relative to the dense baseline. Quantities which decrease from $\mathcal{O}(C)$ to $\mathcal{O}(D)$ during training are denoted by $\mathcal{O}(C)\xrightarrow{t}\mathcal{O}(D)$.}
{\renewcommand{\arraystretch}{1.5} 
\vspace{3mm}

\begin{tabularx}{\textwidth}{@{}ABCDEFGHIJ@{}}
        \toprule
        & \multicolumn{3}{>{\hsize=3.3cm\centering}X}{$\Delta$ Accuracy} & \multicolumn{3}{>{\hsize=4.8cm\centering}X}{\phantom{test}Computational Cost}       & \multicolumn{3}{>{\hsize=3.9cm\centering}X}{Network Size on Device} \\ \cline{1-10} 
$P=$        & 0.01          & 0.001          & 0.0001          & \multicolumn{1}{|c}{At init.} & Training        & Final           & \multicolumn{1}{|c}{At init.}     & Training         & Final    \\ \cline{1-10} 
Random  & -11.9\%           & -66\%              & -66\%               & 0        & $\mathcal{O}(pmn)$          & $\mathcal{O}(pmn)$          & $\mathcal{O}(PN)$        & $\mathcal{O}(PN)$            & $\mathcal{O}(PN)$    \\
IMP     & +0.8\%                 & -7.1\%                  & -64.8\%                   & 0        & $\mathcal{O}(mn)$ $\xrightarrow{t}$ $\mathcal{O}(pmn)$ & $\mathcal{O}(pmn)$          & $\mathcal{O}(N)$         & $\mathcal{O}(N)$ $\xrightarrow{t}$ $\mathcal{O}(PN)$    & $\mathcal{O}(PN)$    \\
FORCE   & -6.6\%            & -26.9\%            & -62.4\%             & $\mathcal{O}(mnk)$   & $\mathcal{O}(pmn)$          & $\mathcal{O}(pmn)$          & $\mathcal{O}(N)$         & $\mathcal{O}(PN)$            & $\mathcal{O}(PN)$    \\
SynFlow & -6.2\%            & -31.6\%            & -60.4\%             & $\mathcal{O}(mnk)$   & $\mathcal{O}(pmn)$          & $\mathcal{O}(pmn)$          & $\mathcal{O}(N)$         & $\mathcal{O}(PN)$            & $\mathcal{O}(PN)$    \\
RigL (ERK) & +0.4\%            & -16.8\%            & -65.7\%             & 0   & $\mathcal{O}(pmn + \frac{1}{\Delta T}mn)$      & $\mathcal{O}(pmn)$          & $\mathcal{O}(PN)$         & $\mathcal{O}(PN)$            & $\mathcal{O}(PN)$    \\
\cline{1-10} 
\textbf{DCTpS}    & -5.8\%            & -15\%              & -22.8\%             & 0        & $\mathcal{O}(q\log q  + pmn)$ & $\mathcal{O}(q\log q  + pmn)$ & $\mathcal{O}(PN)$        & $\mathcal{O}(PN)$            & $\mathcal{O}(PN)$  \\
\bottomrule
\end{tabularx}
}
\label{tab: summary}
\end{table*}

\subsection{Competing Priorities for Sparse Networks}

Traditional pruning algorithms, which prune after or during training, result in a final network with a small storage footprint and fast inference \cite{gale2019state}. However, since these methods initialize networks as dense, and initially train them as such (only to slowly reduce the number of parameters), the overall storage and computational costs of training remain approximately those of a dense network. 

For the benefits of sparsity to extend to training, the network must be pruned before training starts. In \cite{frankle2018the}, and many works since \cite{frankle2020linear, malach2020proving}, researchers have shown the existence of `lottery tickets' -- sparse sub-networks of randomly initialized dense networks, that can be trained on their own from scratch to achieve accuracy similar to that of the full network. This has inspired a surge in recent work on techniques to efficiently prune networks directly at initialization, to identify these trainable, sparse sub-networks. 

Research on prune-at-initialization (PaI) methods has progressed rapidly and achieved impressive test accuracy with well below $1\%$ of the network parameters, see \cite{tanaka2020pruning, jorge2021progressive} and Section \ref{sec:experiments}. However, almost all current PaI algorithms involve the computation of `sensitivity scores' (or a comparable metric) for all candidate parameters in the dense network, which are then used to decide which parameters to prune. Thus, despite a less computationally demanding training procedure, these methods still require the capacity to store, and compute with, the full network on the relevant device (see Table \ref{tab: summary}). 

Ideally, starting with dense networks which are then pruned would be avoided entirely, and only those parameters to be trained would be initialized. The only PaI method to date that can achieve this is random pruning, since it is equivalent to initializing a sparse network with randomly selected sparse support. For high sparsities, however, random pruning achieves significantly lower accuracy than other methods, for details see Section \ref{sec:experiments}.

An alternative approach to training sparse networks from scratch is Dynamic Sparse Training (DST) \cite{mocanu2018scalable, mostafa2019parameter}. In DST methods, the network is initialized as sparse according to some heuristic (thereby also avoiding any initialization and storage of the full network), but the topology - the support set of the non-zero parameters - of the network is updated during training, along with the values of the weights themselves. 

Table \ref{tab: summary} summarises the performance of the current state-of-the-art pruning and DST algorithms in terms of the various competing priorities for sparse networks: accuracy, storage footprint, and computational complexity. 

\subsection{Matching Sparsity vs. Extreme Sparsity}
\cite{frankle2021pruning} note the distinction between what they call `matching sparsities', at which the resulting pruned networks retain (approximately) the same performance as the full dimensional baseline, and `extreme sparsities', at which there is a trade-off between sparsity and performance. Attention is increasingly being paid to the latter regime, which is especially relevant for resource-constrained settings, in which trade-offs may be necessary or considered worthwhile. A crucial question in the extreme-sparsity setting is the rate of accuracy drop-off as sparsity is increased. Prior algorithms like SNIP \cite{lee2018snip} and GraSP \cite{Wang2020Picking} display gradual accuracy decrease up to a point, but then reach a sparsity at which accuracy rapidly collapses to random guessing. The primary improvement achieved by the most recent algorithms, FORCE \cite{jorge2021progressive} and SynFlow \cite{tanaka2020pruning}, is to extend that gradual performance degradation to significantly higher sparsity. The DCTpS method proposed here, too, avoids this `cliff-like' drop-off in accuracy, exhibiting an even more gradual decrease in performance at extreme sparsities, resulting in superior performance in this extreme-sparsity regime. 

\subsection{Contributions}
In this manuscript we introduce a new neural network layer architecture with which networks can be initialized and trained in an extremely low-dimension parameter space. These layers are constructed as the sum of a dense \textit{offset} matrix which need not be stored and has a fast transform, plus a sparse matrix of trainable stored parameters, denoted as DpS to abridge `Dense plus Sparse'.  Consequently, the resulting networks are \textit{in effect} dense, but require the storage of a sparse network with potentially extremely few trainable parameters, and incur the computational cost of very sparse networks. This effective density allows information to continue to propagate through the network at low trainable densities, avoiding unnecessary performance collapse. Our approach and results can be summarised as follows:
\vspace{-1mm}
\begin{itemize}
    \item The neural network layer architectures introduced here are the sum of a discrete cosine transform (DCT) matrix and a sparse matrix, denoted `DCT plus Sparse' (DCTpS). These layers have the same memory footprint as a standard sparse tensor, and a low additional quasi-linear computational overhead above that of sparse layers.
    \item The sparse trainable matrices in all layers are assigned an equal number of trainable parameters, and within each the support is randomly chosen - avoiding any initial storage of, or computation with the dense network. 
    \item A variety of network architectures using these layers are trained to achieve high accuracy, in particular in the extremely sparse regime with weight-matrix density as small as $P=0.0001$, where they significantly outperform prior state-of-the-art methods; for example by up to 37\% on ResNet50 applied to CIFAR100.
    \item Combining DCTpS layers with state-of-the-art Dynamic Sparse Training (DST) techniques yields even further improvements in accuracy at all densities - often in the region of 3\%-7\%.
\end{itemize}

\section{Prior Prune-at-Initialization (PaI) Methods}

Neural network pruning has a large and rapidly growing literature; for wider ranging reviews of neural network pruning see \cite{gale2019state, blalock2020state, liu2020pruning}.  PaI is the subset of pruning research most directly comparable with the `DCT plus Sparse' networks presented here.  For conciseness, we limit our  discussion to the most competitive PaI techniques.  

The most successful PaI methods determine which entries to prune by computing a synaptic saliency score vector \cite{tanaka2020pruning} of the form
\begin{align}
    G(\vw) = \frac{\partial \mathcal{R}}{\partial \vw}\odot \vw, 
\end{align}
where $\mathcal{R}$ is a scalar function, $\vw$ is the vector of network parameters, and $\odot$ denotes the Hadamard product. Those parameters with the lowest scores are pruned.

SNIP \cite{lee2018snip} sets out to prune weights whose removal will minimally affect the training loss at initialization. They suggest `connection sensitivity' as the appropriate metric: $G(\vw) = |\frac{\partial \mathcal{L}(\vw)}{\partial \vw} \odot \vw|$, where $\mathcal{L}$ is the training loss. 

GraSP \cite{Wang2020Picking} instead maximises the gradient norm after pruning. The resulting saliency scores for each parameter are calculated via a Taylor expansion of the gradient norm about the dense initialization, resulting in $G(\vw) = -\left(H \frac{\partial \mathcal{L}(\vw)}{\partial \vw}\right) \odot \vw$, where $H$ is the Hessian of $\mathcal{L}$.

FORCE (and a closely related method, iterative SNIP) \cite{jorge2021progressive}, like GraSP, take into account the interdependence between parameters so as to predict their importance \textit{after pruning}.  They also note, however, that by relying on a Taylor approximation of the gradient norm, GraSP assumes that the pruned network is a small perturbation away from the full network, which is not the case at high sparsities.  Instead they propose letting  $G(\vw) = |\frac{\partial \mathcal{L}(\bar{\vw})}{\partial \vw} \odot \vw|$, where $\bar{\vw}$ is the parameter vector $\vw$ after pruning. They then propose FORCE and Iter-SNIP as iterative algorithms to approximately solve for the score vector $G$ and gradually prune parameters.

SynFlow \cite{tanaka2020pruning} makes use of an alternative objective function $\mathcal{R} = \mathds{1}^\top \left( \prod_{l=1}^L |\vw ^{[l]}|\right)\mathds{1}$, where $|\vw ^{[l]}|$ is the element-wise absolute value of the parameters in the $l^{\text{th}}$ layer, and $\mathds{1}$ is a vector of ones. This allows them to calculate saliency scores $G$ without using any training data.  Like FORCE, their focus extends to extreme sparsities, and their algorithm is designed to avoid layer collapse (pruning whole layers in their entirety) at the highest possible sparsities. Together, FORCE and SynFlow are the current state-of-the-art for pre-training pruning to extreme sparsities.

PaI methods are also sometimes referred to as static sparse training (SST), since the sparse support remains fixed after initialization. DST methods \cite{mocanu2018scalable, mostafa2019parameter}, on the other hand, modify the support set during training, subject to maintaining a fixed sparsity level overall. DST methods are characterised by the criteria according to which they prune some weights and `regrow'/activate others during training, and also whether the layer-wise sparsity levels remain fixed during training, or if connections pruned in one layer can be replaced by additional connections in other layers. The current state-of-the-art for DST is RigL \cite{evci2020rigging}, which maintains a fixed layer-wise sparsity distribution, prunes parameters with the smallest magnitude, and re-activates weights with the largest-magnitude gradients.

Recent work \cite{frankle2021pruning} has shown that given a particular sparsity pattern identified by SNIP, GraSP or SynFlow, one can shuffle the locations of the allotted trainable parameters within each layer, and train the resulting network to matching or even slightly improved accuracy. In other words, they argue, the success of these SST methods is due to their layer-wise distribution of the number of trainable parameters, rather than the particular locations of the trainable parameters within a layer. This somewhat calls into question the role of the proposed saliency metrics used to score the importance of each parameter individually. In contrast, while RigL is also shown to be influenced by the layer-wise sparsity distribution \cite{evci2020rigging}, the premise of DST is precisely that the locations of the trainable parameters within each layer matter fundamentally. Either way, further understanding of, and heuristics for, the ideal layer-wise parameter allocations would be complementary and directly beneficial to the aforementioned PaI and DST methods, as well as `DCT plus Sparse' presented in Section \ref{sec: DCT plus Sparse}.

\section{Restricting Network Weights to Random Subspaces} 

Let $\vw \in \R^n$ denote the full vector of network parameters.  The number of trainable parameters can be reduced by restricting $\vw\in\R^n$ to a $k$-dimensional hyperplane such that
\begin{align}\label{eq: subspace method}
    \vw = \vd + U\theta,
\end{align}
where $\vd$ is an untrainable offset from the origin, $U \in \R^{n\times k}$ is a fixed subspace embedding, and $\theta\in\R^k$ is the vector of $k\ll n$ trainable parameters. A $k$-sparse network ($\vw$ being $k$-sparse, with support set $\mathcal{S}$), such as those generated by PaI methods, represents the specific case when $\vd=0$, and the subspace embedding $U \in \R^{n\times k}$ is a matrix with one nonzero per column and at most one nonzero per row, with their locations determined by $\mathcal{S}$ (we denote this structure for $U$ as `$k$-sparse disjoint').  In this sparse $\vw$ setting, identifying `Lottery Tickets' -- sparse networks (and their initial parameter values) which can be trained to high accuracy from scratch -- can be viewed as identifying the appropriate $U$ and $\theta_0$.

The model \eqref{eq: subspace method} was explored in \cite{li2018measuring} where they showed it is possible to \textit{randomly} draw the offset $\vd$ and subspace $U$, and to retain accuracy comparable to that of the full $n$-dimensional network by training only the $k\ll n$ parameters in $\theta$.  In their work, $\vd$ is drawn from a traditional, say Gaussian, initialization known to have desirable training properties, and $U$ has geometry-preserving properties similar to drawing $U$ uniformly from the Grassmannian; for details see \cite{li2018measuring} S7. The smallest possible dimension $k$ for which such subspace training achieved 90\% of the accuracy of a dense network was termed the `intrinsic dimension' of the loss surface, as the ability to successfully train a network in a random low-dimensional subspace indicates some low-dimensional structure in the loss landscape.

\begin{figure}[h!]
    \centering
    \includegraphics[width=0.45\textwidth]{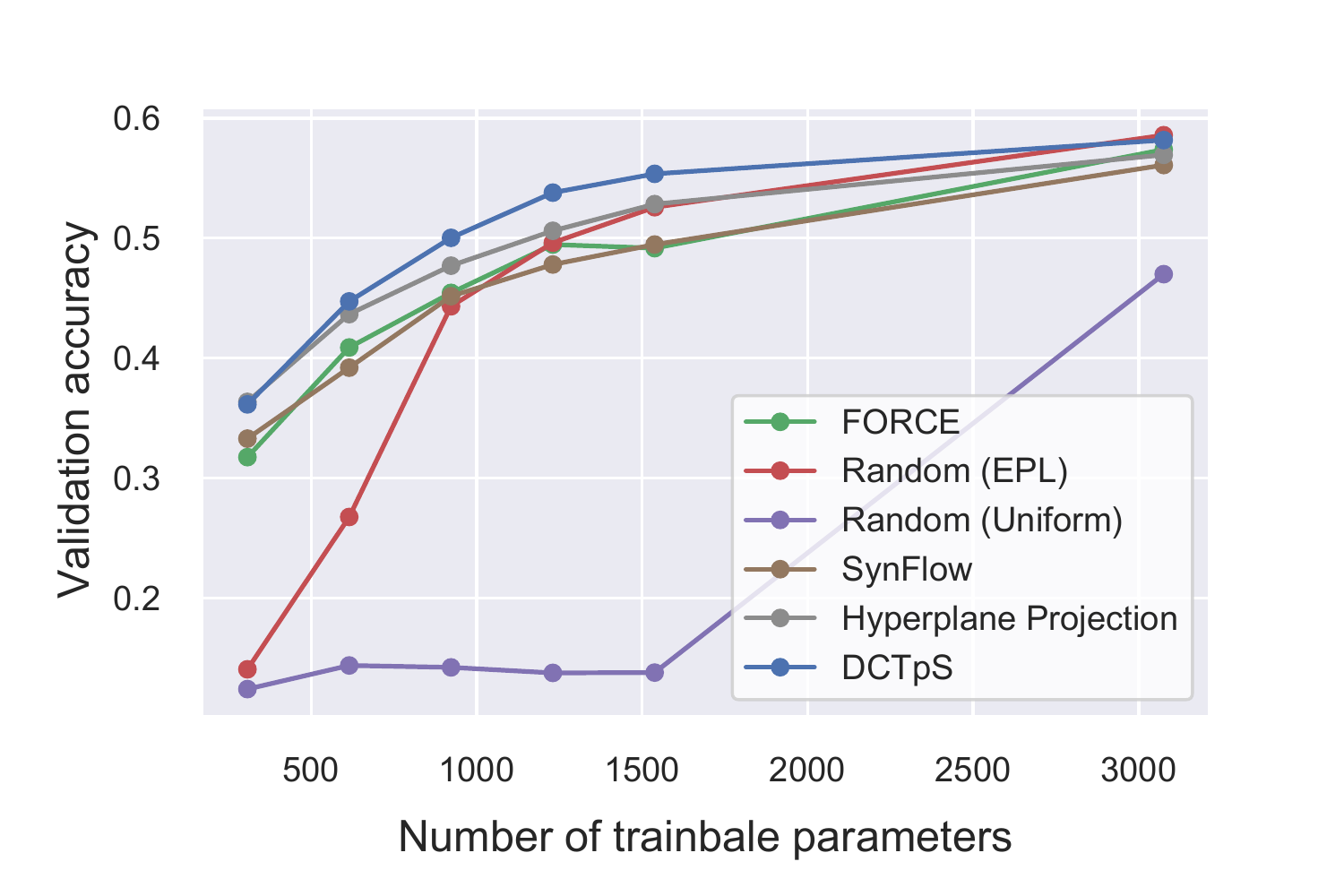}
    \caption{Different subspace selection methods applied to Lenet-5 \cite{lecun1998gradient} and trained on CIFAR10. We report maximum validation accuracy, averaged over three runs, at each subspace dimension (number of trainable parameters).}
    \label{fig:lenet}
\end{figure}

In Figure \ref{fig:lenet}, we repeat one of the experiments from \cite{li2018measuring}, comparing their method, which we denote as `Hyperplane Projection', with random pruning and the aforementioned PaI methods, on Lenet-5 with CIFAR10.  The performance of Li et al.'s method stands in stark contrast with the performance of random pruning at initialization, which corresponds to \eqref{eq: subspace method} with $\vd=0$ and $U$ being $k$-sparse disjoint, with $\mathcal{S}$ selected uniformly at random.  Despite both methods involving training in randomly selected subspaces, `Hyperplane Projection' far outperforms random pruning at $k\ll n$. Furthermore, in this low-dimensional regime, `Hyperplane Projection' even outperforms state-of-the-art PaI algorithms.

However, in the context of PaI algorithms it is important to note that despite having the same number of trainable parameters $k$, the networks based on the `Hyperplane Projection' model \eqref{eq: subspace method}, with $\vd \in \R^n$ dense, do not afford any memory and computational benefits over a dense network.  

In the following section we propose an alternative subspace model to \eqref{eq: subspace method}, `DCT plus Sparse', which combines the benefit of the dense nonzero offset $\vd$ of \cite{li2018measuring} with the specially structured sparse $U$ as used in PaI methods, without needing to store the offset $\vd$. Moreover, we show that state-of-the-art test accuracy is obtained even while selecting the location of the 1-sparse rows in $U$ according to a simple random equal-per-layer heuristic, avoiding the initial calculation of parameter saliency scores.

\section{DCT plus Sparse (DCTpS) Network Layers}\label{sec: DCT plus Sparse}

The network parameters $\vw$ in \eqref{eq: subspace method} which PaI methods sparsify are typically only the network weight matrices as they usually comprise the largest number of trainable parameters\footnote{Tanaka et al. briefly extend their analysis to batchnorm layers in their Appendix (Section 10) \cite{tanaka2020pruning}.}.  In the specific case we consider, combining $\vd$ dense and non-trainable, with $U$ being $k$-sparse disjoint, the associated weight matrices $W \in \R^{m\times n}$ comprising $\vw$ in  \eqref{eq: subspace method} can be expressed as $W = D + S$, where $D$ is dense, but fixed (i.e. non-trainable $\vd$), and $S$ is sparse, with fixed sparse support (i.e. non-trainable $U$) and trainable values within that support (corresponding to the trainable $\theta$).  As $D$ is dense, the sparse matrix $S$ can be initialized as zero, and the training of $S$ corresponds to adjusting only $k\ll n$ entries within $D$.  To allow for an additional bulk scaling of $D$ by a trainable parameter\footnote{We note that the inclusion of an $\alpha$ scaling parameter for $D$ is a departure from a standard subspace training model since it enables the re-scaling of different sections of $\vd$ independently during training, but it adds expressive power with almost no overhead. See the Supp. Mat. for experiments without an $\alpha$ parameter.} $\alpha$, similar to batch normalization \cite{ioffe2015batch}, we consider $W = \alpha D + S$.

In order to maintain the low network size on device and to reduce the computational burden of applying $W$ with a dense  component (and $W^\top$ in the backward pass), we treat the dense offset $D$ as the action of the discrete cosine transform (DCT) matrix\footnote{If the input dimension is less or greater than the output dimension, we zero-pad the input or truncate the output respectively.} resulting in 
\begin{align}\label{eqn:wx full}
    Wx &= \alpha \text{DCT}(x) + Sx, \\
    W^\top v &= \alpha \text{DCT}^{-1}(v) + S^\top v.
\end{align}
The DCT can be applied in near linear, $q\log{q}$, computational cost (where $q = \max(m,n)$), and need not be directly stored. Consequently, this layer architecture \eqref{eqn:wx full} retains the benefit of $W$ being dense, while having the on device storage footprint of a sparse network and at the minimal overhead of requiring an additional computational $q\log q$ cost. We refer to  layers parameterized by~\eqref{eqn:wx full} as `DCT plus Sparse' (DCTpS) layers. There are, of course, many other candidate matrices for $D$ with the same or similar properties, which together constitute a more general `Dense plus Sparse' (DpS) layer class, but we restrict our attention to DCTpS in this paper, deferring alternative choices of fast transforms to later investigation.

The framework \eqref{eqn:wx full} applies equally to convolutional layers. Each step in a convolution can be cast as a matrix-vector product $Wx$, where $x \in \R^\text{patch\_dim}$ is a vectorised `patch' of the layer input, and $W \in \R^{\text{out\_channels}\times \text{patch\_dim}}$ has the filters as its rows. Back-propagation through a 2D convolutional layer involves convolutions with (rotated versions of) the layer's filters, each step of which can be implemented as $W\hat{v}$, where $\hat{v}$ is a permuted version of a patch $v$. See the Supplementary Material for more details.

The state-of-the-art PaI algorithms, such as FORCE and SynFlow, require initially storing and computing with  densely initialized weight matrices so as to compute saliency scores for each parameter.  To avoid these costs, we use only a simple heuristic to determine the locations of the trainable parameters in DCTpS networks: we allocate an equal number of trainable parameters to each layer, and select their locations within each weight tensor uniformly at random.

This `Equal per layer' (EPL) heuristic achieves the basic goal of maintaining some amount of trainability within each layer, but is otherwise naive. While we will show that even something as simple as EPL is sufficient for state-of-the-art results with DCTpS networks, there is likely scope for improved heuristics for the allocation of trainable parameters, which - as noted in \cite{frankle2021pruning} - may be the most relevant feature of a PaI method, and thus may further improve performance. In the Supplementary Material, we include experiments with another naive heuristic which distributes parameters evenly across filters, rather than layers, and achieve similar results.

\section{Experiments}\label{sec:experiments}

SynFlow \cite{tanaka2020pruning} and FORCE \cite{jorge2021progressive} have recently emerged as the state-of-the-art PaI algorithms in the extreme-sparsity regime, significantly outperforming the prior state-of-the-art methods SNIP and GraSP. We thus focus our experiments on comparing DCTpS with SynFlow and FORCE\footnote{A revised version of  \cite{jorge2021progressive} includes a closely related variant of FORCE, called Iter-SNIP with very similar results. It suffices to compare our method to FORCE.}, as well as RigL (ERK), the current state-of-the-art DST method. A full description of all experimental setups and hyperparameters is included in the Supplementary Material, and our code is available at \href{https://github.com/IlanPrice/DCTpS}{github.com/IlanPrice/DCTpS}. For all plots in this section, solid lines represent test accuracy averaged over three runs, and shaded regions (though often too small to make out) represent the standard deviation. The dashed black lines denote the unpruned dense network baseline accuracy, while dashed colored lines indicates where an algorithm breaks down and is thus unable to prune the network to the specified sparsity.

\noindent \textbf{Combining DCTpS with RigL:} Though DCTpS networks have been presented in the framework of static sparse training, they can straightforwardly be combined with DST methods like RigL, which are then used to iteratively update the support set of the DCTpS layers' sparse $S$ matrices. We test this in the experiments below, with positive results.

\begin{figure*}[!h]
    \centering
    \includegraphics[width = 0.85\textwidth]{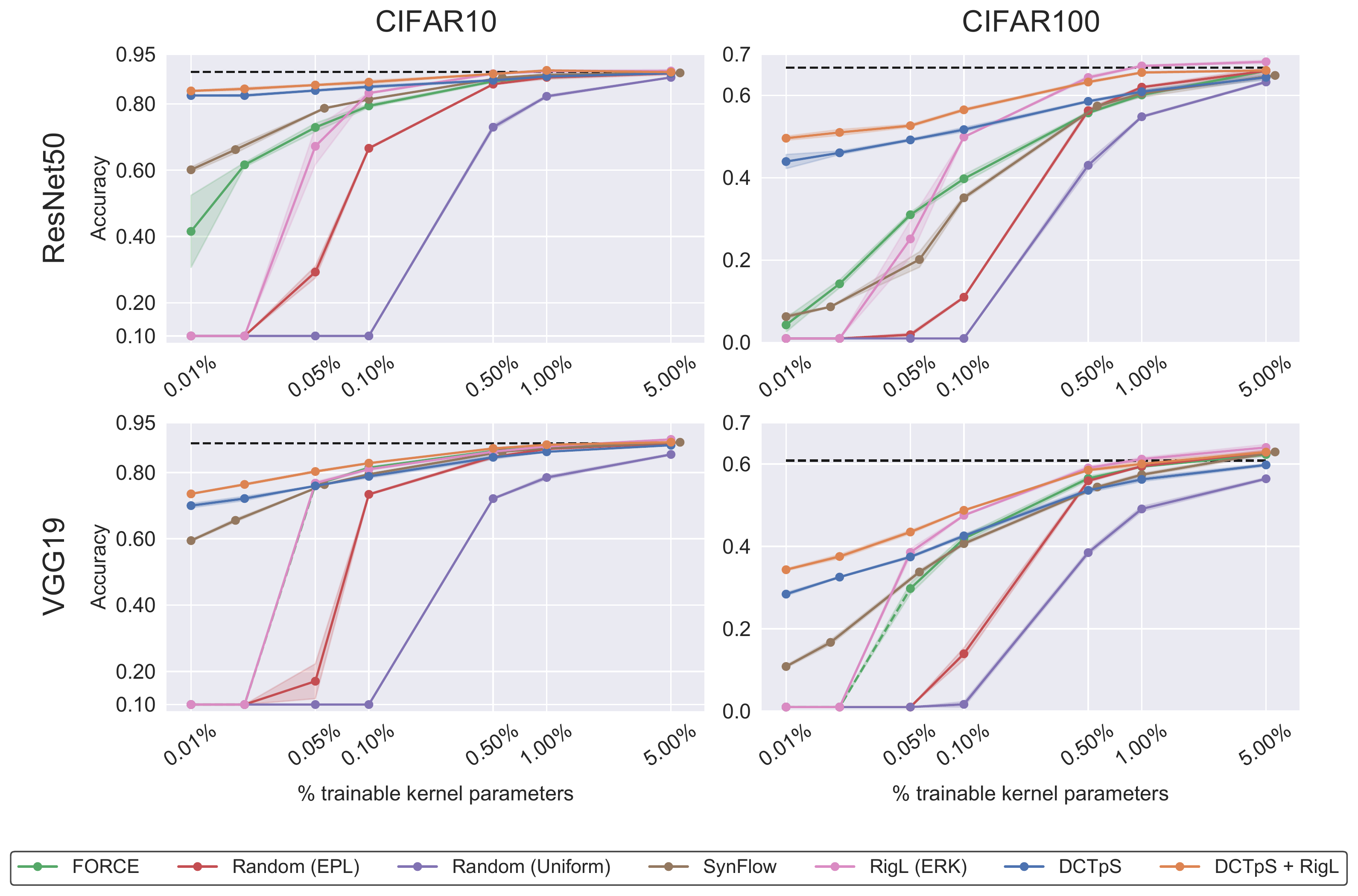}
    \caption{Test accuracy on CIFAR10 and CIFAR100 datasets using sparse ResNet50 and VGG19 architectures. DCT plus Sparse (DCTpS) networks (with EPL parameter allocation) are compared with FORCE, SynFlow, RigL, and random pruning methods.}
    \label{fig:vgg_and_resnet50}
\end{figure*}

\noindent \textbf{Random Pruning Comparisons:} As noted above, the support sets of trainable nonzero parameters in `DCTpS' networks are selected without any calculations involving the full network. Relevant SST comparisons in this respect are thus variants of random pruning, since initializing sparse matrices is equivalent to initializing them as dense and pruning randomly. Globally uniform random pruning, which we denote as Random (uniform) in Figures \ref{fig:lenet} - \ref{fig:jacobian spectra}; is often included as a baseline in works such as \cite{jorge2021progressive, lee2018snip, Wang2020Picking}.  We include an additional random sparse initialization, Random (EPL), which uses the same heuristic for distributing trainable parameters as we use for DCTpS networks (with the difference being that the trainable entries in the sparse weight matrices are not initialized as 0, but according to a standard initialization scheme). We note that this Random (EPL) heuristic significantly outperforms Random (uniform) and other heuristics (see Supp. Mat.), and even matches state-of-the-art pruning methods for densities greater than $\sim 1\%$, see e.g. Figure~\ref{fig:vgg_and_resnet50}.

\subsection{Lenet-5}

We first consider the small Lenet-5 architecture on CIFAR10 so as to compare sparse network methods against the `Hyperplane Projection' method of \cite{li2018measuring}, which is computationally demanding despite having few trainable parameters, due to the nature of its chosen random subspace.  Figure \ref{fig:lenet} illustrates that Hyperplane Projection achieves validation accuracy superior to all PaI methods tested except DCTpS which matches or exceeds its accuracy.  The efficacy of the Hyperplane Projection method helps illustrate the value of the affine offset in \eqref{eq: subspace method}, while the even superior accuracy of DCTpS shows that the offset can be deterministic and the hyperplane sparse and axis-aligned as in \eqref{eqn:wx full}.

\subsection{ResNet50 and VGG19 on CIFAR10 and CIFAR100}

ResNet50 and VGG19 are selected as the primary architectures to benchmark the PaI methods considered here; this follows \cite{jorge2021progressive} and allows direct comparison with related experiments conducted therein. 

Figure \ref{fig:vgg_and_resnet50} displays the test accuracy of these architectures, applied to CIFAR10 and CIFAR100 datasets, as a function of the percentage of trainable parameters within the weight matrices determined by the aforementioned PaI algorithms. 
At $5\%$ density all PaI algorithms are able to obtain test accuracy approximately equal to that of a dense network.  Random (uniform) and Random (EPL) initializations exhibit a collapse or significant drop in accuracy once the density drops below $1\%$ and $0.5\%$ respectively, but at greater densities they roughly match or even outperform other PaI methods. Below $0.5\%$, DCTpS has superior or equal test accuracy compared to both SynFlow\footnote{We note that one feature of SynFlow's saliency scores is that they grow very large for large networks, and thus in order to successfully prune ResNet50 with SynFlow it is necessary to switch from float32 to float64 to avoid overflow.} and FORCE.  The superior accuracy of DCTpS is most pronounced as the density decreases to $0.01\%$; for example, in the case of ResNet applied to CIFAR10, DCTpS achieves accuracy $22\%$ above the next most effective method SynFlow, and moreover retains accuracy in excess of $80\%$. With the introduction of dynamic sparsity, RigL achieves superior accuracy to all static sparse approaches at higher densities (ranging from $5\%$ to $0.5\%$), after which it is overtaken by DCTpS. Applying RigL to the sparse matrices in DCTpS networks yields significant improvements over static sparse DCTpS networks; for example, considering ResNet50 applied to CIFAR100, DCTpS with RigL achieves $\approx 50\%$ accuracy at $0.01\%$ density, as compared with $44\%$ with static DCTpS. DCTpS combined with RigL achieves accuracy comparable to or better than all methods, including RigL, at densities of $1\%$ or below, again with the most significant margins at the most extreme sparsities.

The test accuracy of all methods are somewhat lower for VGG19 applied to CIFAR10 and CIFAR100.  Up to approximately $0.05\%$ density, each of SynFlow, FORCE, and DCTpS achieve accuracy  within approximately $2\%$ of each other.  FORCE fails to generate a network for densities below $0.05\%$, denoted by dashed green lines, while the benefits of DCTpS over SynFlow become apparent for these smaller densities where its test accuracy is approximately $10\%$ and $20\%$ greater on CIFAR10 and CIFAR100 respectively. Once again RigL outperforms static sparse approaches at densities ranging from $5\%$ to $0.1\%$, but collapses below this level. Combining RigL with DCTpS is comparable to or better than all other methods at all densities, and significantly outperforms static DCTpS (by 3\% - 6\% at all densities on CIFAR100, for example).

\begin{figure*}[!h]
    \centering
    \includegraphics[width = 0.85\textwidth]{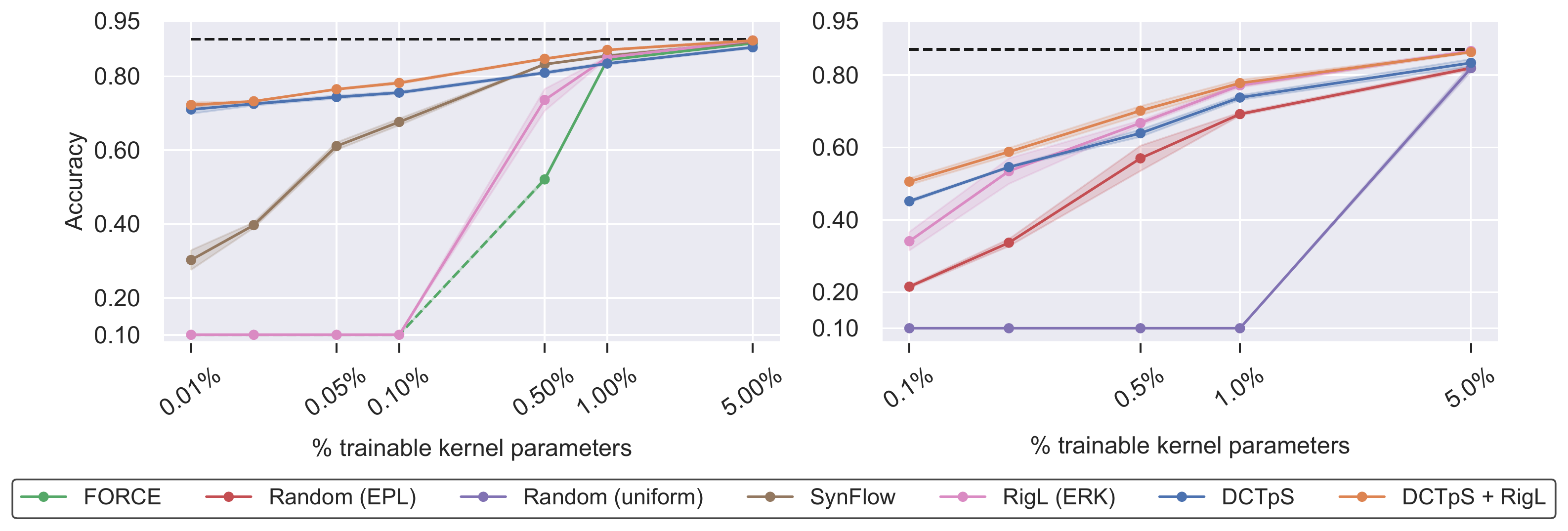}
    \caption{\textit{Left:} Comparison of DCTpS against FORCE, SynFlow and RigL on CIFAR10 with MobileNetV2 \cite{sandler2018mobilenetv2}. \textit{Right:} Comparison of DCTpS with RigL and Random Pruning methods on CIFAR10 with FixupResNet110 \cite{zhang2018residual}.}
    \label{fig:mobilenet and fixupresnet}
\end{figure*}

One important point to underscore is that in these experiments, as is typical in prior PaI experiments \cite{jorge2021progressive}, the reported sparsity refers only to the weights in linear and convolutional layers. Large networks like ResNet50 and VGG19 also have bias and batchnorm parameters (making up 0.22\% and  0.06\% of trainable parameters of the respective architectures). That these are not prunable in our experiments imposes a floor on the overall number of trainable parameters remaining in the network. The plateau in performance exhibited by DCTpS networks in Figure \ref{fig:vgg_and_resnet50} is thus testament to their ability to preserve information flow through the network despite extremely few trainable weights, thereby preserving the capacity of the network endowed by other remaining trainable parameters.

Shrinking the storage footprints of batchnorm networks to their most extreme limits will require the development and incorporation of methods to prune batchnorm parameters as well. Such methods can be incorporated into DCTpS networks (and other PaI methods).

\subsection{MobileNetV2 and Fixup-ResNet110}
Next, DCTpS is compared to other PaI algorithms and RigL on two architectures which are less overparameterized than ResNet50 and VGG19. First, we consider MobileNetV2 \cite{sandler2018mobilenetv2}, originally proposed as a PaI test-case in \cite{jorge2021progressive}, which has approximately $10\%$ of the parameters in ResNet50.  Figure~\ref{fig:mobilenet and fixupresnet} (left) shows test accuracy for MobileNetV2 applied to CIFAR10, demonstrating trends similar to those in Figure~\ref{fig:vgg_and_resnet50}, with DCTpS exhibiting superior performance as sparsity increases, retaining approximately $70\%$ accuracy with as few as $0.01\%$ of the networks weights. RigL's accuracy drops significantly below 1\% density, but combining DCTpS with RigL achieves the best accuracy of all methods at all densities.

\begin{figure*}[!h]
    \centering
    \includegraphics[width=\textwidth]{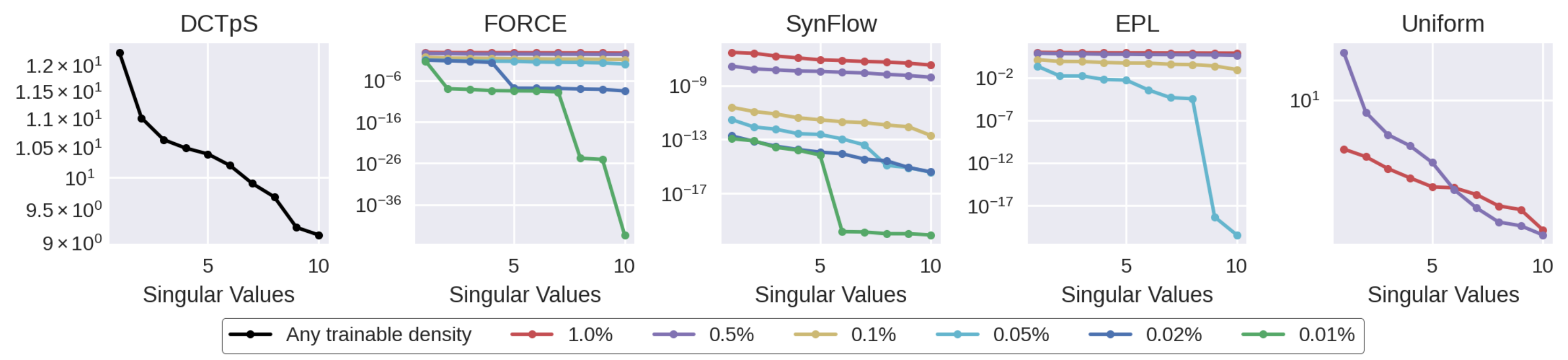}
    \caption{Spectrum of the Jacobian of Resnet50 on CIFAR10, pruned to different sparsities with varies methods, at initialization. If a curve does not appear in a given plot, it means that the spectrum was identically zero for that density. The DCTpS plot shows only one curve since its Jacobian does not depend on the number of trainable weights.}
    \label{fig:jacobian spectra}
\end{figure*}

Second, we include experiments for Fixup-ResNet110 \cite{zhang2018residual} on CIFAR10. `Fixup' ResNets were developed in order to enable the efficient training of very deep residual networks without batchnorm, to the same accuracy as similarly sized batchnorm networks. In Fixup-ResNet110, all but 282 of its 1720138 parameters are `prunable', practically eliminating the overall density floor caused by batchnorm parameters in the other large networks considered in this section. 

The Fixup initialization involves initializing some layers to zero (in particular the classification layer and the final layer in each residual block), as well as re-scaling the weight layers inside the residual branches. Since our layers are initialized as DCTs, we mimic these effects by setting the $\alpha$ parameter to 0 or to the appropriate scaling factor. We use the code provided by the authors\footnote{\href{https://github.com/hongyi-zhang/Fixup}{https://github.com/hongyi-zhang/Fixup}} to obtain the baseline and random pruning results, and create DCTpS Fixup ResNets by simply replacing the Linear and Convolutional layers with their corresponding DCTpS variants, and initializing appropriately.

Figure \ref{fig:mobilenet and fixupresnet} (right) illustrates the test accuracy of DCTpS, RigL, Random (EPL) and Random (uniform).  FORCE and SynFlow cannot be directly applied to FixupResNet110 as they both assign a saliency score of 0 to all parameters and consequently have no basis on which to select which entries to prune.  At 0.1\% density, with only 2000 trainable parameters in total, spread across 110 layers, DCTpS retains approximately $45\%$ test accuracy, outperforming the Random (EPL) by more than $20\%$. DCTpS outperforms RigL below $0.5\%$ density, and DCTpS with RigL matches or outperforms all other methods at all densities.

\begin{figure}[!h]
    \centering
    \includegraphics[width = 0.33\textwidth]{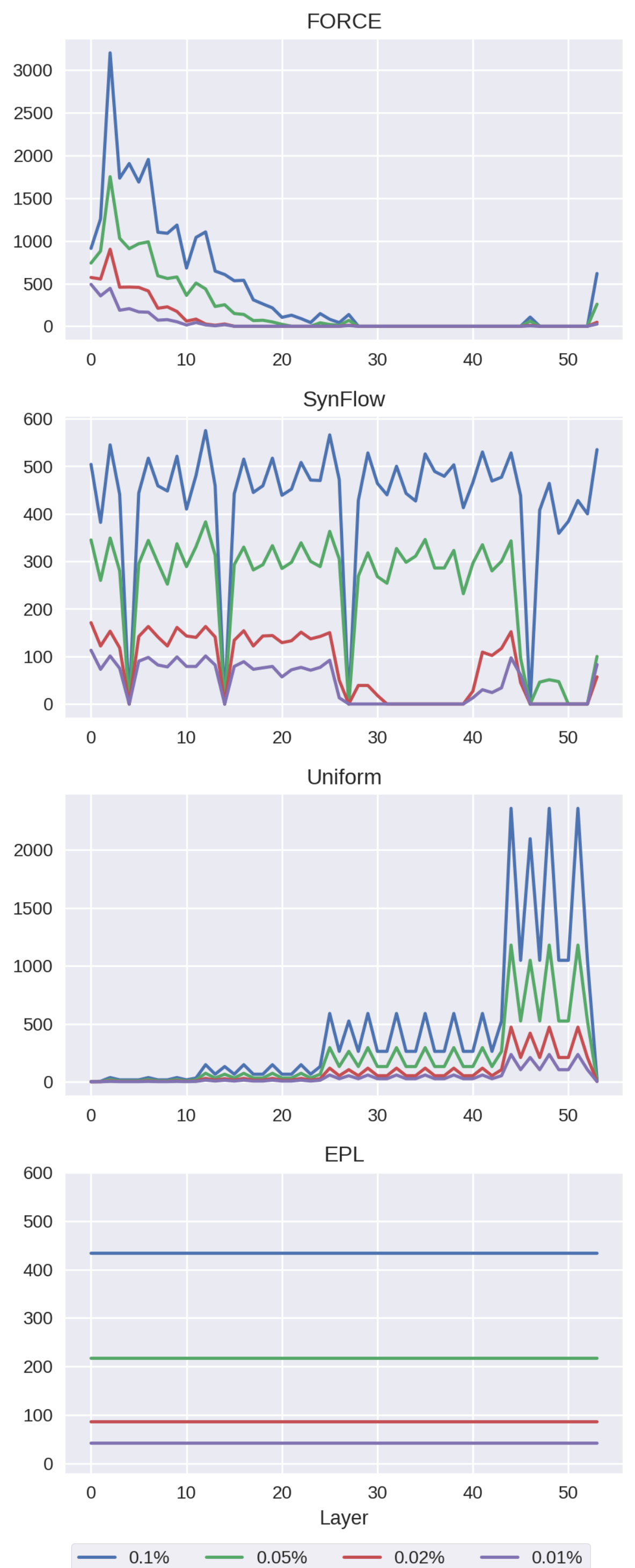}
    \caption{Number of remaining parameters (y-axis) in the prunable layers (x-axis) in ResNet50 after pruning with different PaI methods at a variety of densities. Each color curve represents a different global density. The insets zoom in on portions of the plot with the lowest totals, to illustrate where methods do or do not prune weight tensors in their entirety.}
    \label{fig:layer non_zeros}
\end{figure}

\subsection{Run Times and Theoretical Complexity}\label{sec: complexity}

The operations of both fully connected and convolutional layers can be framed in terms of matrix multiplication, and thus we may discuss the complexity of a DCTpS layer as compared with a standard sparse layer by considering a matrix-vector product $Wx$, where $W\in \R^{m \times n}$. If $Wx$ is parameterized as in Equation \ref{eqn:wx full}, with $S$ containing $pmn$ non-zeros, then theoretically the storage cost of $W$ is just the storage cost of $S$,  $\mathcal{O}(pmn)$, while the computational cost of applying the layer (computing $Wx$) becomes $\mathcal{O}(q\log q + pmn)$, where $q=\max(m,n)$. We note here that while storage requirements can decrease to arbitrarily low levels, depending on $p$, once $p<\mathcal{O}(\log q / r)$ where $r=\min(m,n)$ , further sparsification results in only minimal computational savings.

However, current implementations of deep learning packages render these computational and storage gains purely theoretical, for now at least, and thus plots showing run-times and storage costs corresponding to the above expressions are not included. Firstly, most standard deep learning libraries are not optimised for sparse tensor operations, which affects the realisation of the potential benefits of all PaI techniques, as well as our DCTpS approach. Secondly, efficient DCTpS networks would require optimising the fast transforms in these packages, and appropriately building in their auto-differentiation.

\section{Spectral Analysis and Distribution of Nonzeros}

Figure \ref{fig:vgg_and_resnet50} shows that the  accuracy of Random (uniform) collapses to random guessing as the density decreases from 0.5\% to 0.1\%, and for Random (EPL) this occurs at 0.02\%.  In both cases the percentage of trainable parameters at which the pruned network becomes un-trainable are those at which the spectrum of their Jacobian ($J_{ij} = \frac{\partial f_i}{\partial x_j}$) becomes equal to 0 at initialization. This can be seen in Figure \ref{fig:jacobian spectra}, which displays the associated leading singular values of the Jacobian in the exemplar case of ResNet50 applied to CIFAR10, for different PaI methods at varying densities.  As previously mentioned, FORCE and SynFlow do not reduce to random guessing at any sparsity tested with ResNet50, and correspondingly there is no sparsity at which the Jacobian spectrum fully collapses to zero.  It appears that the primary factor for trainability of PaI networks is determined by whether the spectrum is or is not 0, as opposed to the scale of the spectrum -- SynFlow remains competitive with FORCE even at densities for which SynFlow's Jacobian has a largest singular value of approximately $10^{-13}$ whereas the corresponding singular value of FORCE is  $>10^{10}$ greater.

Since DCTpS networks are always, in effect, dense networks, including at initialization, the spectrum of the Jacobian, shown in Figure \ref{fig:jacobian spectra}, does not depend on the sparsity of its trainable weights, and thus does not collapse even as the number of trainable parameters approaches zero. This likely relates to their ability to be trained even with extremely sparse and randomly distributed trainable weights.

It has been alternatively conjectured that these noted collapses in accuracy occur due to `layer collapse', where one or more layers have \textit{all} of their parameters set to zero, as was noted in \cite{tanaka2020pruning} for SNIP and GraSP. However, in residual networks, due to the existence of multiple branches through which information may flow, it is possible to prune multiple weights tensors in their entirety without collapsing performance. Indeed this phenomenon is observed in Figure \ref{fig:layer non_zeros}, which includes plots of the total number of trainable parameters per layer determined by different PaI methods at different sparsities, applied to ResNet50 on CIFAR10. Comparing Figure \ref{fig:layer non_zeros} to the test accuracy of the corresponding experiments in Figure \ref{fig:vgg_and_resnet50}, shows that pruning all parameters in one or more weight tensors is neither necessary nor sufficient for accuracy collapse. Neither SynFlow nor FORCE exhibit complete performance collapse at any density, despite both methods pruning multiple layers completely.  Conversely, Figure \ref{fig:vgg_and_resnet50} shows Random (EPL) on ResNet50 reduces to random guessing at a density of $0.02\%$ or less, despite the fact that each single layer contains approximately 100 trainable parameters.

The distribution of nonzeros generated by PaI methods was recently investigated in depth by \cite{frankle2021pruning}; and in particular, the value of carefully selecting the locations of trainable parameters with PaI methods.  It was observed that given a particular sparsity pattern identified by PaI methods, the location of aa layer's nonzeros can be shuffled and the resulting network can be trained to similar or even improved accuracy, suggesting the success of a PaI algorithm may be determined primarily by the number of trainable parameters per layer rather than which entries within the layers are selected.  Figure \ref{fig:layer non_zeros} illustrates that SynFlow, for densities down to $0.1\%$, allocates trainable parameters approximately equally per layer (with 4 notable exceptions, which turn out to correspond to those shortcut connections which were prunable). Yet despite this rough similarity in distribution at $0.1\%$ density, we observe substantially different test accuracy for SynFlow and Random~(EPL).

\section{Conclusions and Further Work}
In this work we have shown that adding a layer-wise offset to sparse subspace training significantly improves test accuracy in the extreme-sparsity regime.  DCT plus Sparse (DCTpS) layers provide an elegant way of achieving this offset with no extra storage cost, and only a small computational overhead.  Moreover we have shown that simple heuristics can be used to randomly select the support sets for their sparse trainable weight tensors, avoiding any initial storage of, or computation with the full network. DCTpS networks achieve state-of-the-art results at extreme trainable sparsities, and are competitive with the state-of-the-art at lower sparsities. These results are further improved when DST methods like RigL are applied to DCTpS networks.

There are numerous clear avenues for extending and complementing this research. As noted, research on better heuristics or other ways to choose the trainable sparse support in the DCTpS layers may improve performance beyond our simple EPL heuristic. Moreover, this should be combined with research on pruning or removing batchnorm parameters, since DCTpS layers only enable the pruning of trainable weight tensors. Furthermore, there may well be an optimal initialization of the $\alpha$ parameter in DCTpS layers, and it may not need to be trained.  As the scaling parameter of the network weight tensors, this research would be analogous to work on the optimal scale parameter to use when initializing Gaussian weights, of which there is plenty \cite{glorot2010understanding, he2015delving, xiao2018dynamical}. Finally, the use of other fast deterministic transforms, or other even more efficient ways to implement a dense offset, may yield further improvements.

\section*{Acknowledgements}
This publication is based on work supported by the Alan Turing Institute under the EPSRC grant EP/N510129/1.

\bibliography{paper}
\bibliographystyle{icml2021}

\appendix

\renewcommand\thefigure{\thesection.\arabic{figure}} 
\renewcommand\thetable{\thesection.\arabic{table}} 

\section{Additional Experiments}\label{app: experiments}

\subsection{Tiny Imagenet}
\begin{figure}[h!]
    \centering
    \includegraphics[width=0.45\textwidth]{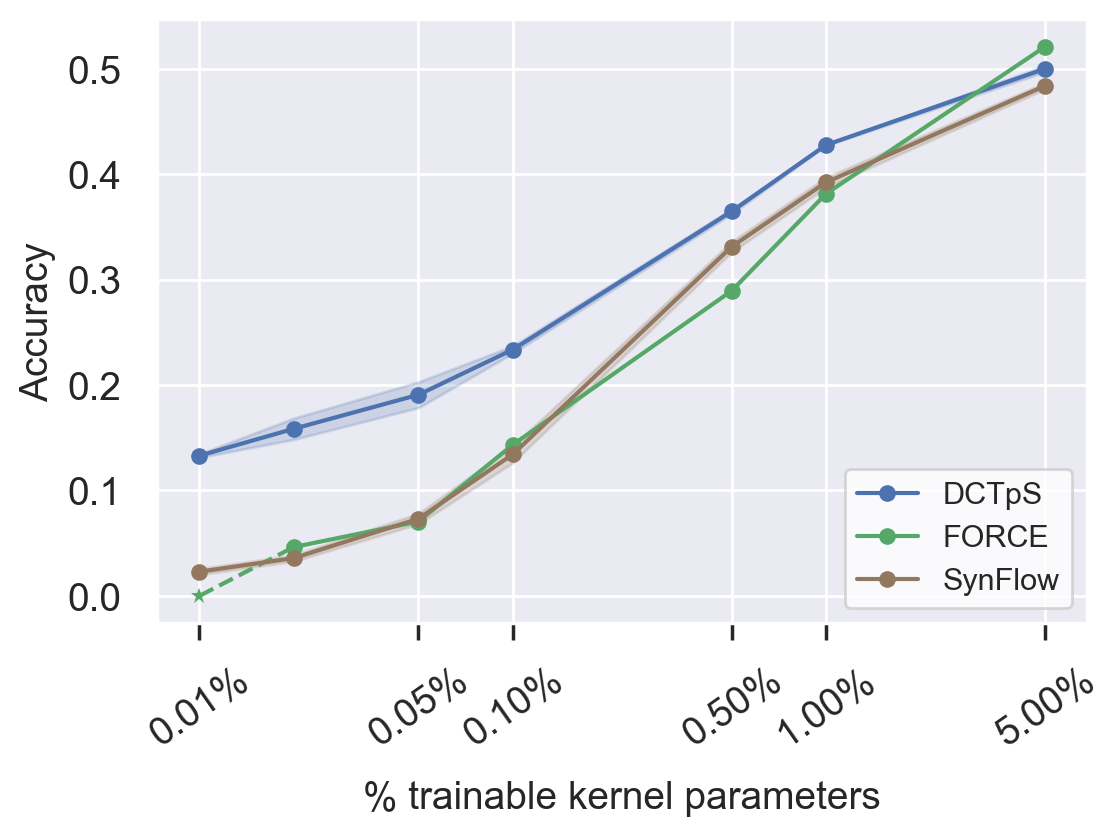}
    \caption{Comparing DCTpS with FORCE and SynFlow on Tiny Imagenet with ResNet18.}
    \label{fig: TI r18}
\end{figure}

Figure \ref{fig: TI r18} compares the performance of DCTpS with SynFlow and FORCE\footnote{We note that FORCE was particularly unstable in these experiments, failing to prune the network to the required density in at least one of its three runs, at every density less than 0.5\%. In these cases, accuracy is averaged over only those runs in which FORCE succeeded. At 0.01\% density, FORCE failed on all three runs.} on Tiny Imagenet with ResNet18.  DCTpS obtains higher validation accuracy than both FORCE and SynFlow for all densities less than or equal to 1\%, and by 0.1\% density DCTpS is outperforming them by approximately 10\% accuracy. This confirms that the superior accuracy of DCTpS networks at low densities is maintained when the difficulty of the problem is scaled up.

\vspace{-3mm}
\subsection{Equal-per-filter (EPF) Support Distribution}

All DCTpS experiments in Section 5 used the EPL heuristic to distribute trainable parameters between layers. Another naive heuristic which achieves the basic goal of maintaining some trainablility throughout the network is an `Equal per Filter' (EPF) approach: given a specified sparsity, the total number of trainable parameters for the whole network is calculated, and divided equally between all convolutional filters (or rows in the case of linear layers). Within each filter, the locations of those trainable parameters is chosen uniformly at random. 

\begin{figure}[h!]
    \centering
    \includegraphics[width=0.45\textwidth]{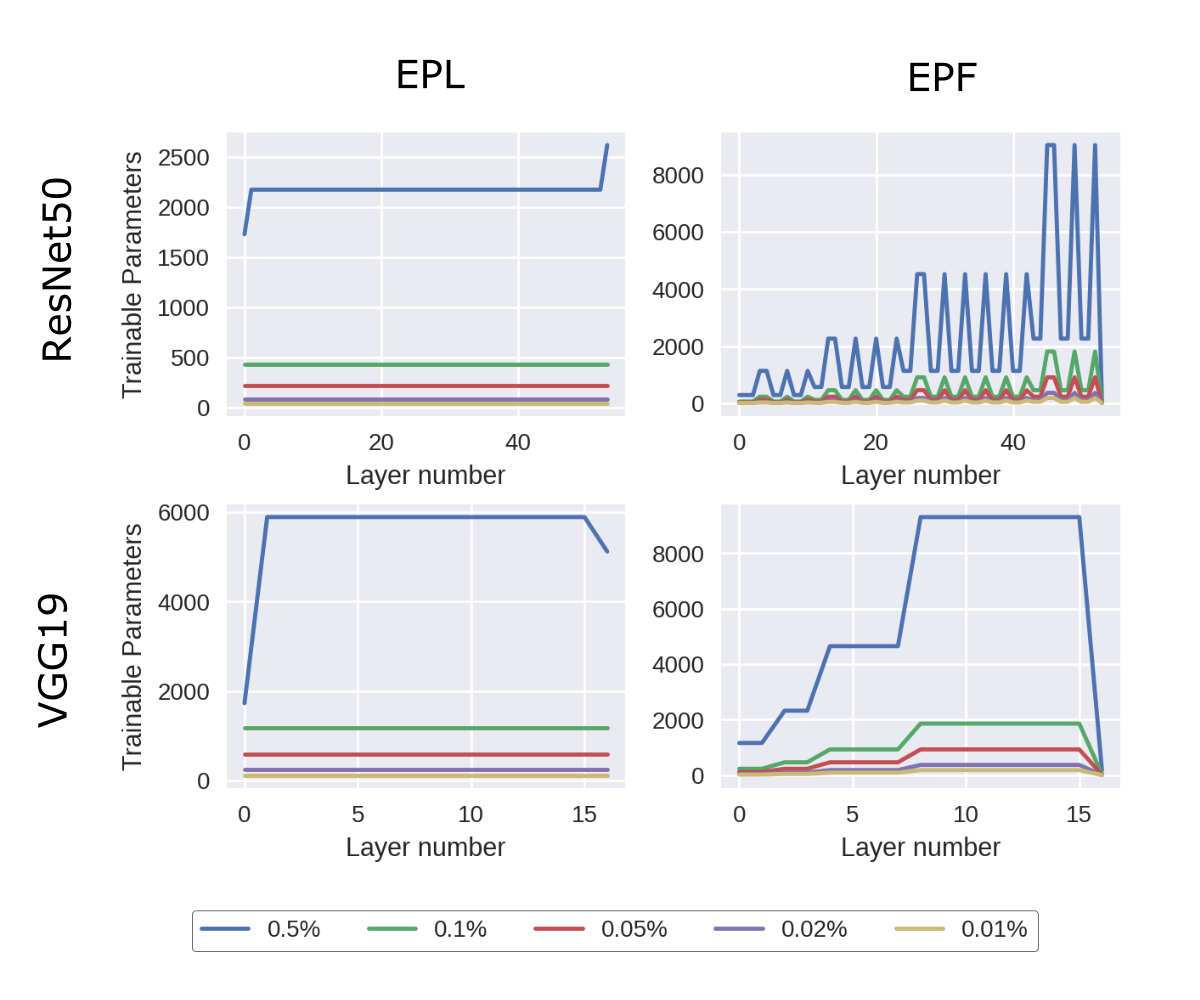}
    \caption{The total number of trainable parameters per prunable layer determined by EPL and EPF heuristics, on ResNet50 and VGG19 with 10 output classes.}
    \label{fig:EPL vs EPF nonzeros}
\end{figure}

\begin{figure*}[h!]
    \centering
    \includegraphics[width=0.8\textwidth]{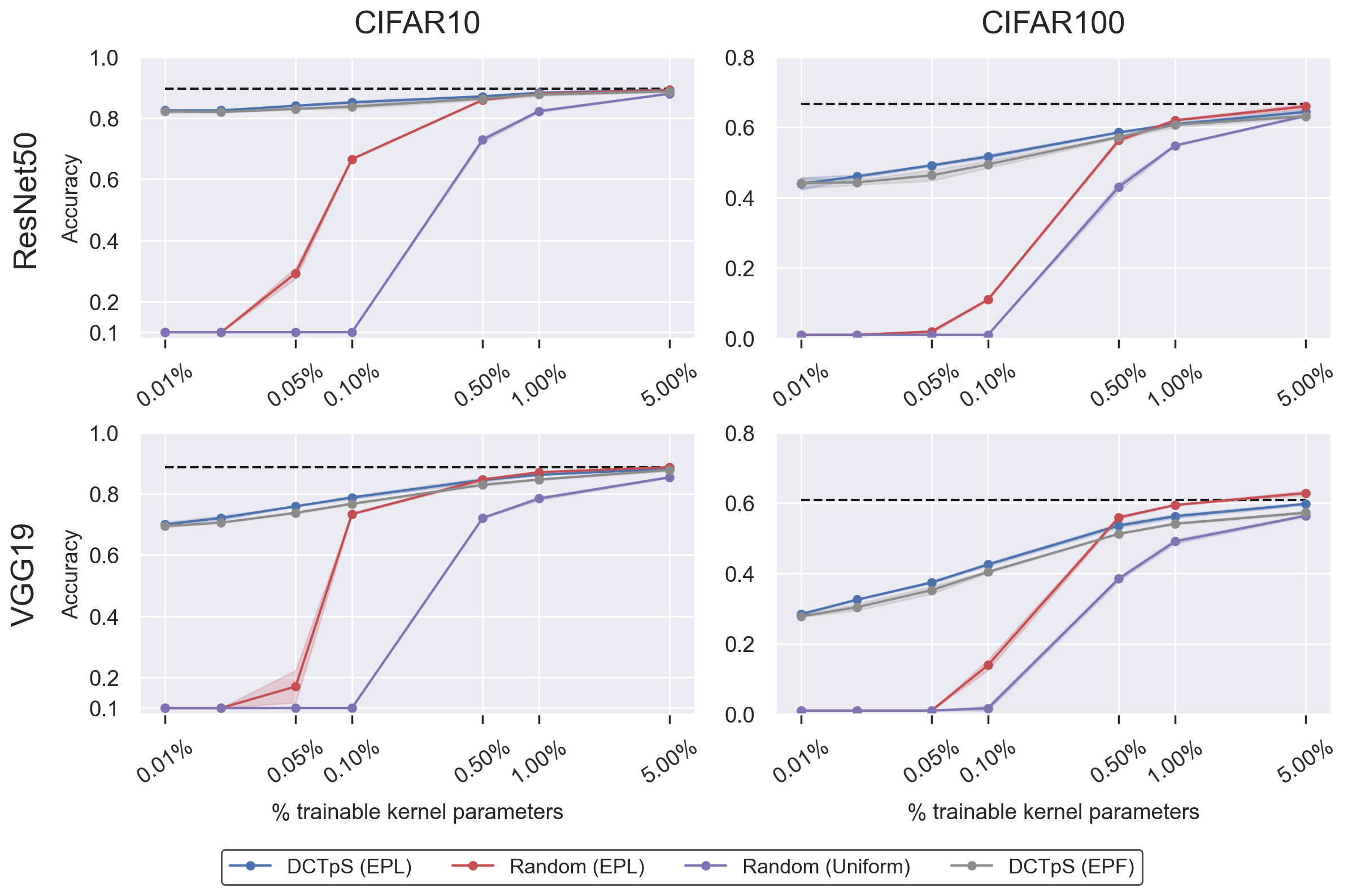}
    \caption{Comparative accuracy of EPF and EPL heuristics for DCTpS networks.}
    \label{fig:EPL vs EPF}
\end{figure*}

Figure \ref{fig:EPL vs EPF nonzeros} highlights that the EPL and EPF heuristics result in substantially and qualitatively different layer-wise parameter allocations. Nevertheless, Figure \ref{fig:EPL vs EPF} shows that both methods achieve very similar accuracy across all tested densities, though EPF consistently performs marginally worse. This observation lends further support to the hypothesis that, provided a suitable offset $\vd$ is used, there is a large class of subspace embeddings from which it suffices to draw $U$ randomly to achieve high accuracy, and in particular that this class includes $k$-sparse disjoint $U$ with a variety of support distributions.

\subsection{Training with SGD as Opposed to Adam} \label{app: sgd}
The best test accuracy for large networks like ResNet and VGG is typically obtained by using SGD with momentum and a specified learning rate schedule, rather than adaptive methods like Adam. However, the results obtained with SGD are sensitive to hyperparameters like the initial learning rate and the learning rate schedule. Adam, though it tends to achieve lower final accuracy, is less sensitive to these hyperparameters. This makes Adam a sensible training algorithm for experiments in which the goal is to compare the relative drop in accuracy caused by one or other pruning method, as opposed to a goal of achieving maximum possible accuracy overall. Thus Adam is used, with default settings, as the training algorithm for the experiments presented in Section 5. In order to preserve comparison with prior work, and to illustrate that our results are not unique to Adam, additional experiments using SGD with momentum on ResNet50, VGG19, MobileNetV2 and FixUpResNet110 are included here. A single, course sweep of base learning rates [0.1, 0.07, 0.05, 0.03, 0.01] was done with a DCTpS ResNet50 applied to CIFAR100, at 1\% density, to select a base learning rate of 0.03, which was then used to train all DCTpS architectures, at all densities, without further fine-tuning. For PaI on standard architectures, a base learning rate of 0.1 was used as done in prior work \cite{jorge2021progressive}.

\begin{figure*}[!h]
    \centering
    \includegraphics[width=0.8\textwidth]{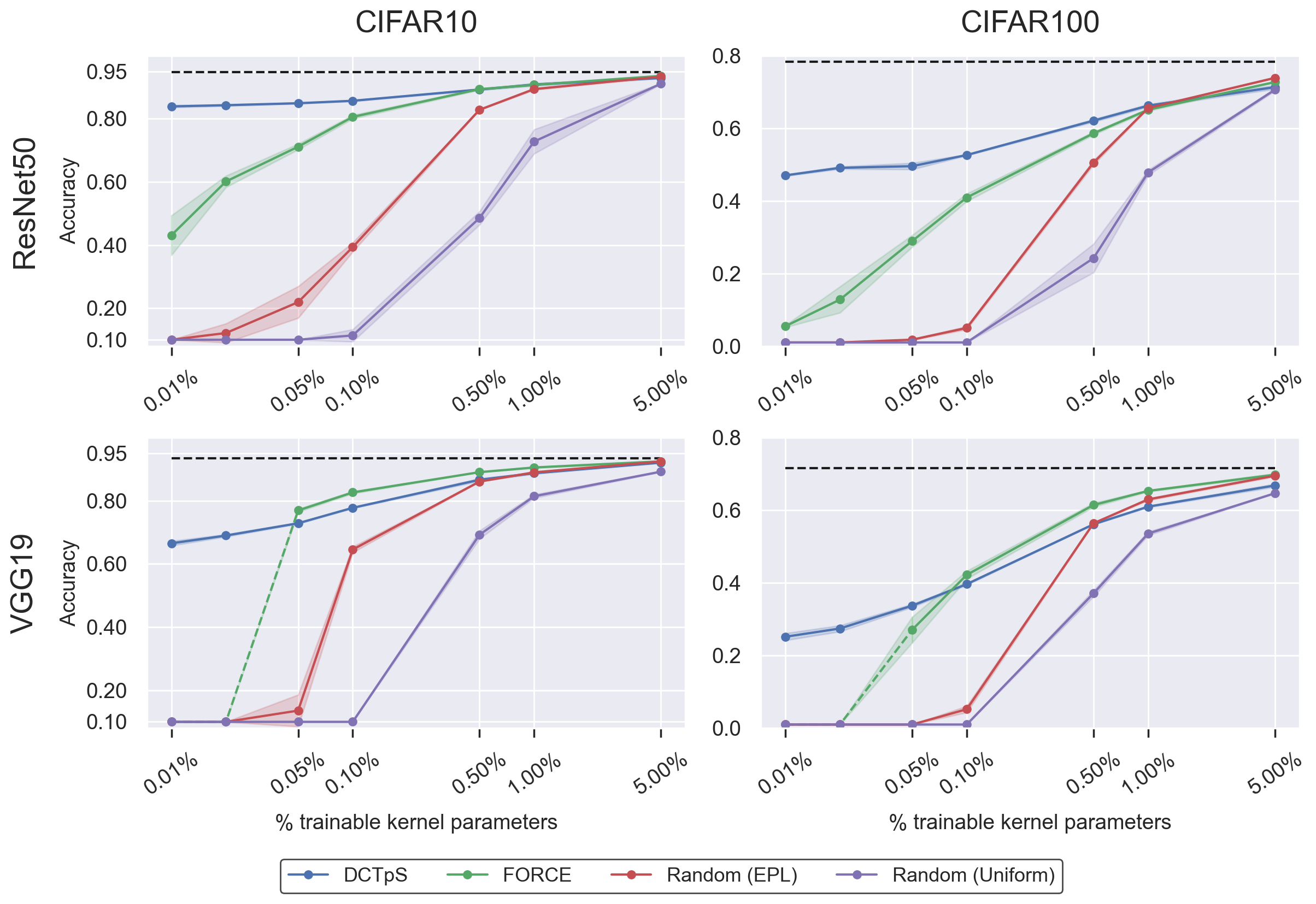}
    \caption{Experiments training ResNet50 and VGG19 with SGD (with momentum), on CIFAR10 and CIFAR100.}
    \label{fig:vgg and resnet with sgd}
\end{figure*}

\begin{figure*}[!h]
    \centering
    \includegraphics[width=0.8\textwidth]{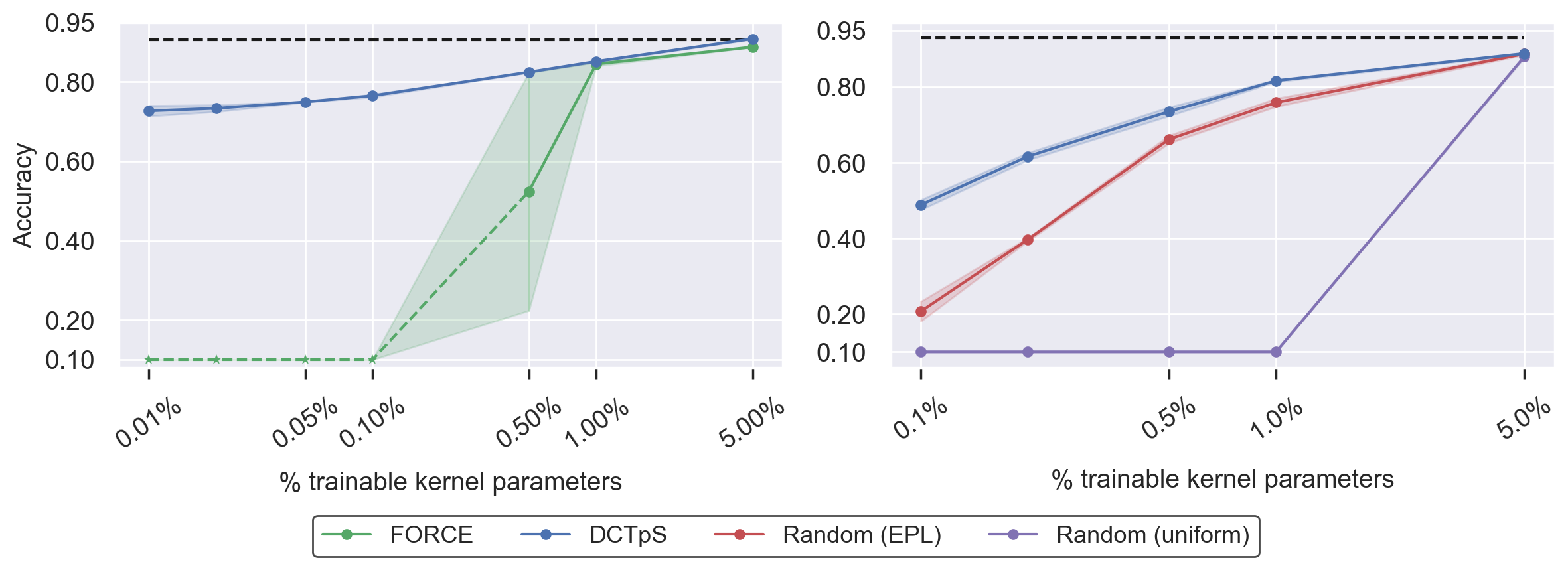}
    \caption{Experiments training MobileNetV2 (left) and FixupResnet110 (right) with SGD (with momentum), on CIFAR10.}
    \label{fig:mobilenet and fixupresnet with sgd}
\end{figure*}

Test accuracy is shown in Figure \ref{fig:vgg and resnet with sgd} for ResNet50 and VGG19, and Figure \ref{fig:mobilenet and fixupresnet with sgd} for MobileNetV2 and FixupResNet110\footnote{SynFlow was not able to be included in these additional supplementary experiments having only recently been published \cite{tanaka2020pruning}.}. The results exhibit qualitatively similar trends to those observed in Section 5's  Figures 2 and 3, though with slightly higher overall accuracy, in particular at higher densities, as expected. 

\subsection{Fixed $\alpha$}
\begin{figure*}[!h]
    \centering
    \includegraphics[width=0.8\textwidth]{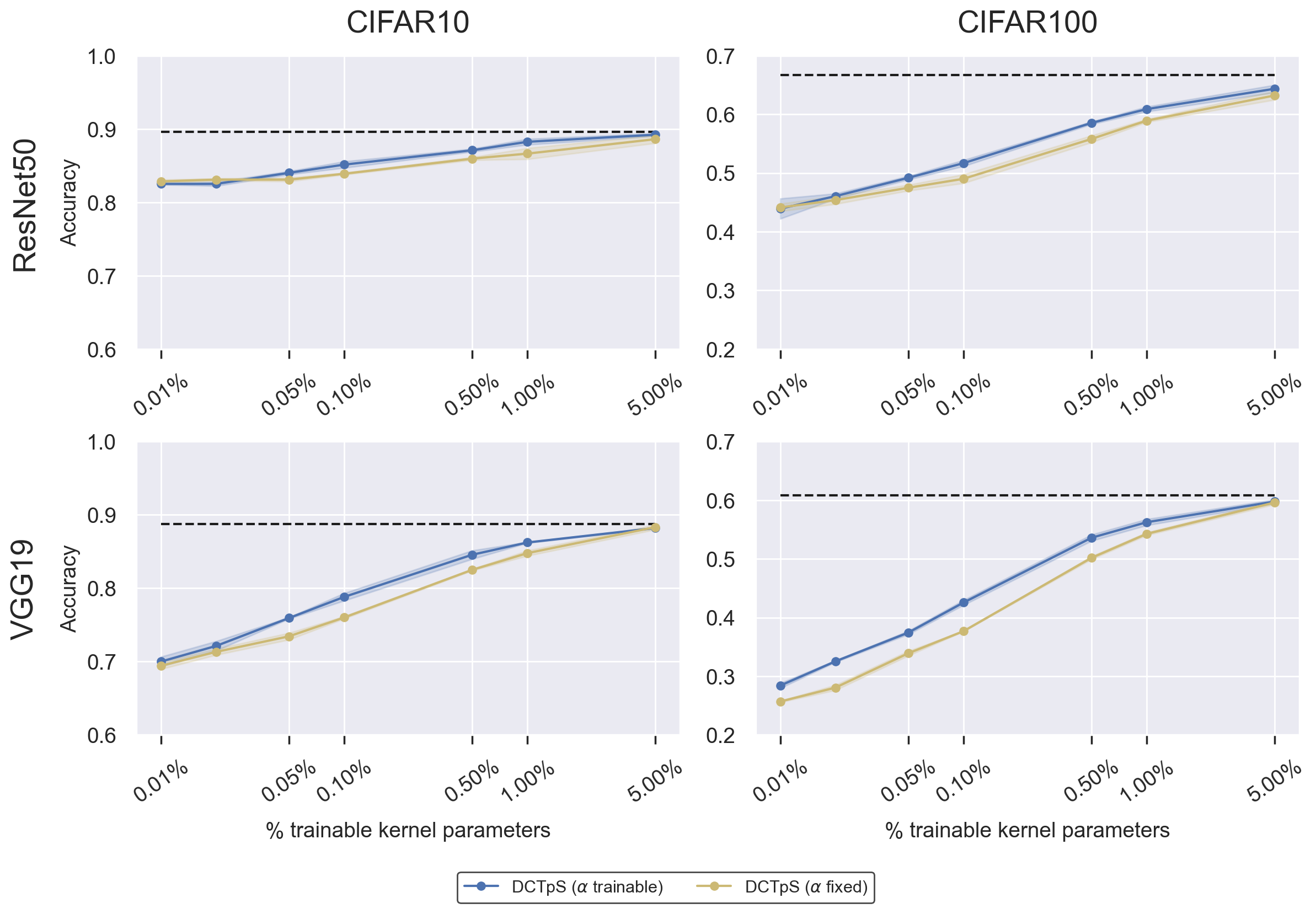}
    \caption{Comparison of letting $\alpha$ be a trainable parameter with fixing $\alpha = 1$ throughout training in DCTpS networks.}
    \label{fig:fixed alpha}
\end{figure*}

In Figure \ref{fig:fixed alpha} we explore the impact of allowing $\alpha$ to be trainable, as compared with being fixed at $\alpha=1$. The latter case corresponds exactly to subspace training, since the offset $\vd$ and embedding $U$ are then fixed during training. We can see that even with a fixed $\alpha$, the same general trends are observed. However, we consistently achieve a 2\%-3\% increase in accuracy by allowing $\alpha$ to be trainable. We conjecture that with the appropriate initialisation of $\alpha$, this gap would disappear, and $\alpha$ would not need to be trained.

\subsection{Comparing Layer-wise Parameter Allocation Heuristics}
\begin{figure*}[!h]
    \centering
    \includegraphics[width=0.8\textwidth]{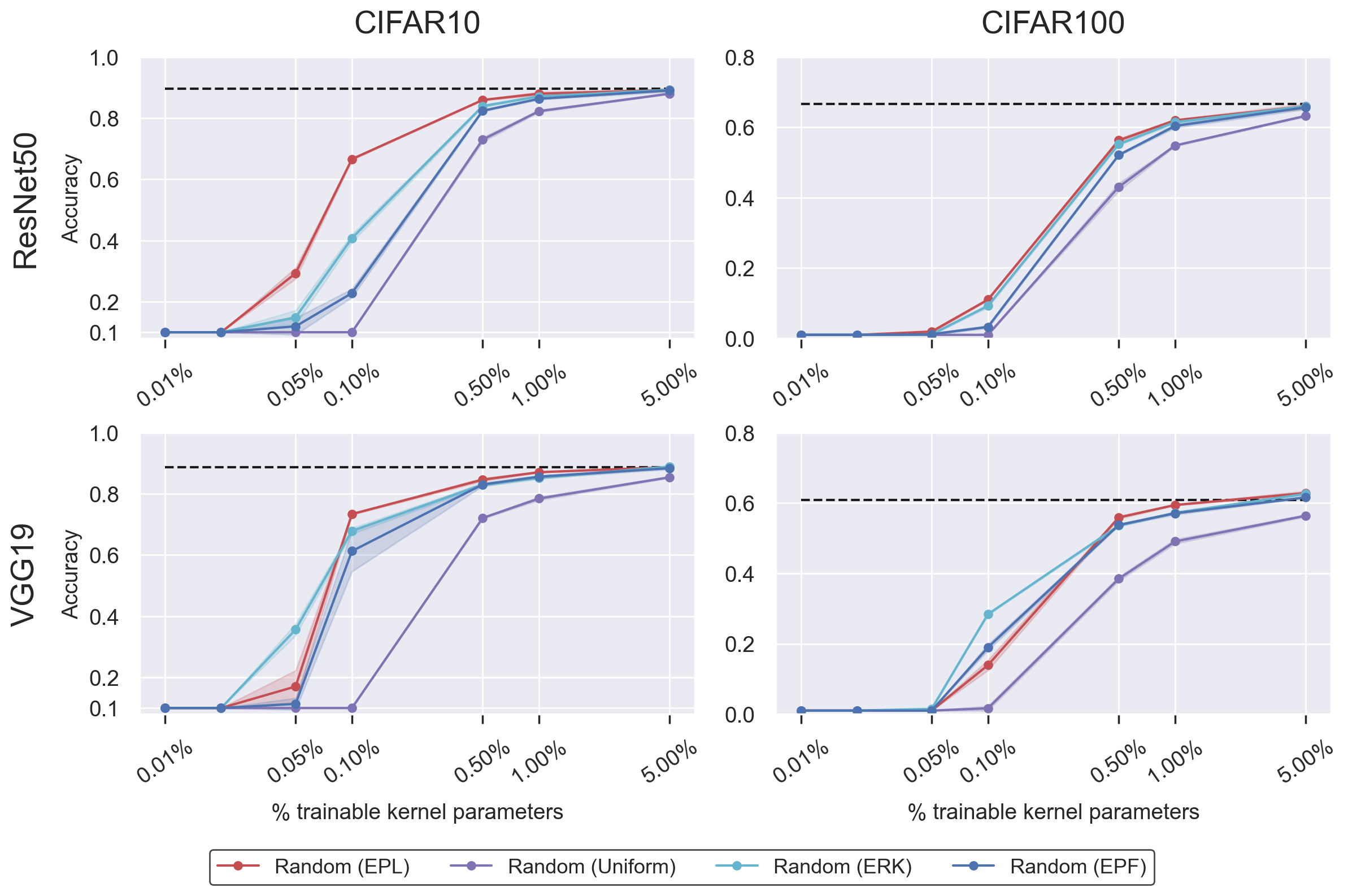}
    \caption{Comparison of different heuristics for distributing trainable parameters between network layers.}
    \label{fig:nonzero distributions}
\end{figure*}

In Figure \ref{fig:nonzero distributions} we compare different heuristics for distributing trainable parameters between network layers -- in particular, uniform density per layer (uniform), equal number of parameters per layer (EPL), equal number of parameters per filter (EPF) and the ERK distribution used in \cite{evci2020rigging}. Though no heuristic performs best in all cases at all densities, EPL performs best in the majority of network-density-dataset combinations - suggesting it could be a useful heuristic for future works to try in methods with fixed layer-wise sparsity distributions, and should certainly be included as an alternative baseline to uniform random pruning in future works.

\section{DCTpS Implementation}
In the code used to run the experiments in this paper, the DCT components of DCTpS layers have been implemented by setting the tensor $W$ to be the DCT matrix (matrix of DCT basis vectors), as opposed to implementing them via fast transforms. This is because deep learning libraries, as currently implemented, are optimised for the former rather than the latter.

\noindent \textbf{Linear Layers:} To be precise, in Linear layers with input $\in \R^n$ and output $\in \R^m$, with $q = \max (m,n)$, the DCT matrix $W \in \R^{q \times q}$ is formed and then truncated to size $m \times n$ by removing the surplus right-most columns (if $m>n$) or bottom-most rows (if $m<n$). Multiplication by this matrix is equivalent to a DCT, with a zero-padded input if $m>n$, or a truncated output if $m<n$.  

\noindent \textbf{Convolutional Layers:}. In a convolutional layer, with $m$ $(k \times k \times n)$ filters, $q = \max(k^2n, m)$, the DCT matrix $W \in \R^{q \times q}$ is formed and truncated to size $k^2 n \times m$ and reshape appropriately. In the accompanying code, a simple test script is provided to confirm that our implementation indeed computes the DCT of each patch.

\begin{figure}[h!]
 \centering
 \includegraphics[width=0.45\textwidth]{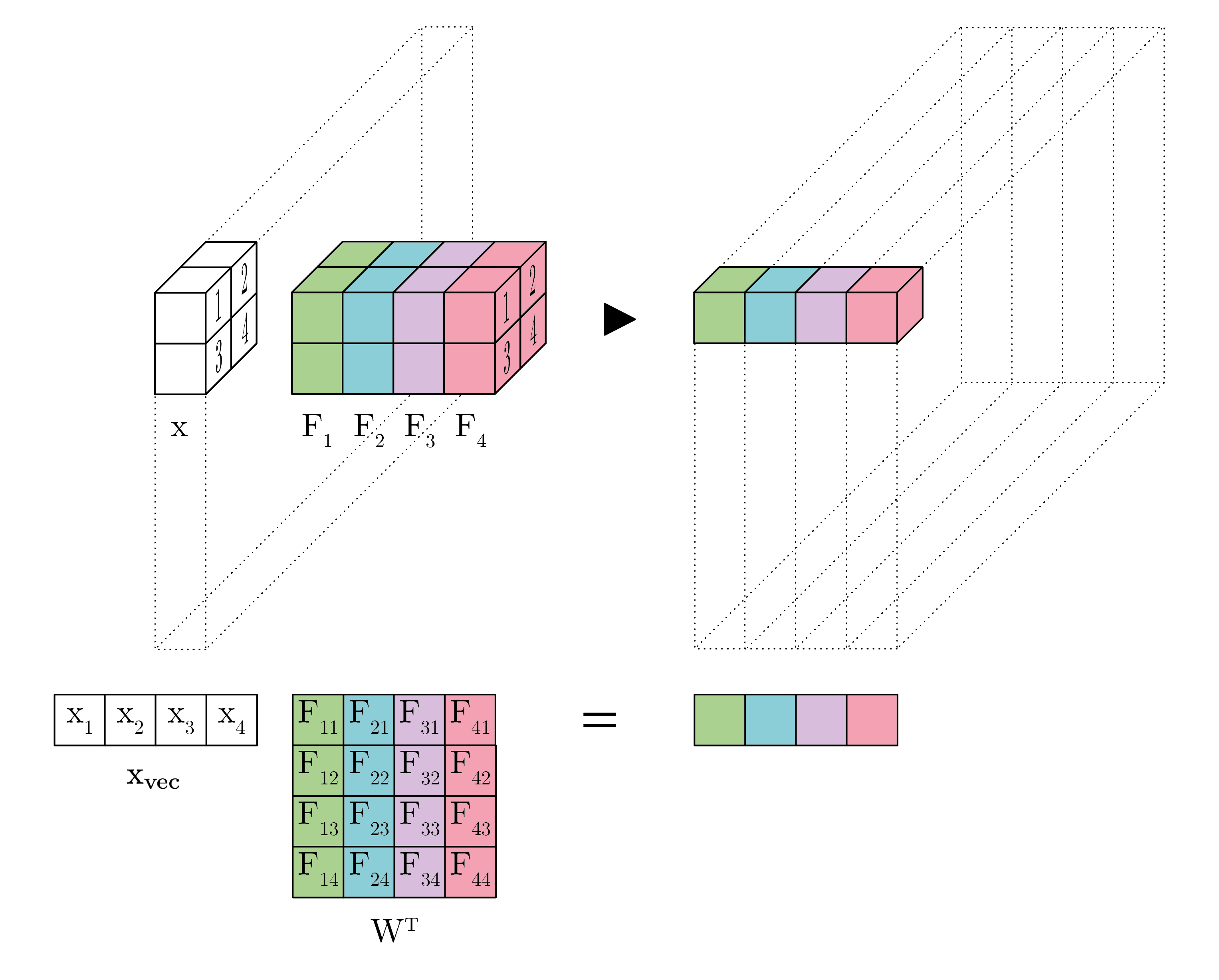}
 \caption{Illustration of the matrix-vector product involved in each step of a 2D convolutional layer. Computing the DCT of each patch is equivalent to taking $W$ to be a DCT matrix.}
 \label{fig:dctps_conv}
\end{figure}

Figure \ref{fig:dctps_conv} provides a simple visualisation of how this is compatible with the efficiency of DCTpS layers, if implemented correctly. In the forward pass, each step of the convolution involves taking a patch of the image, and - for each filter - computing the sum of the elementwise product of the patch and the filter. Flattening (commonly known as `lowering') the filters and the input patch, this is equivalent to a matrix-vector product (and indeed convolutional layers are commonly implemented with this `lower $\rightarrow$ matrix multiply $\rightarrow$ lift' approach \cite{hadjis2015caffe}). The DCT part of DCTpS convolutional layers set this flattened matrix of filters to be the DCT matrix, and is thus equivalent to computing the DCT of each patch.

\begin{figure}[h!]
     \centering
     \includegraphics[width=0.45\textwidth]{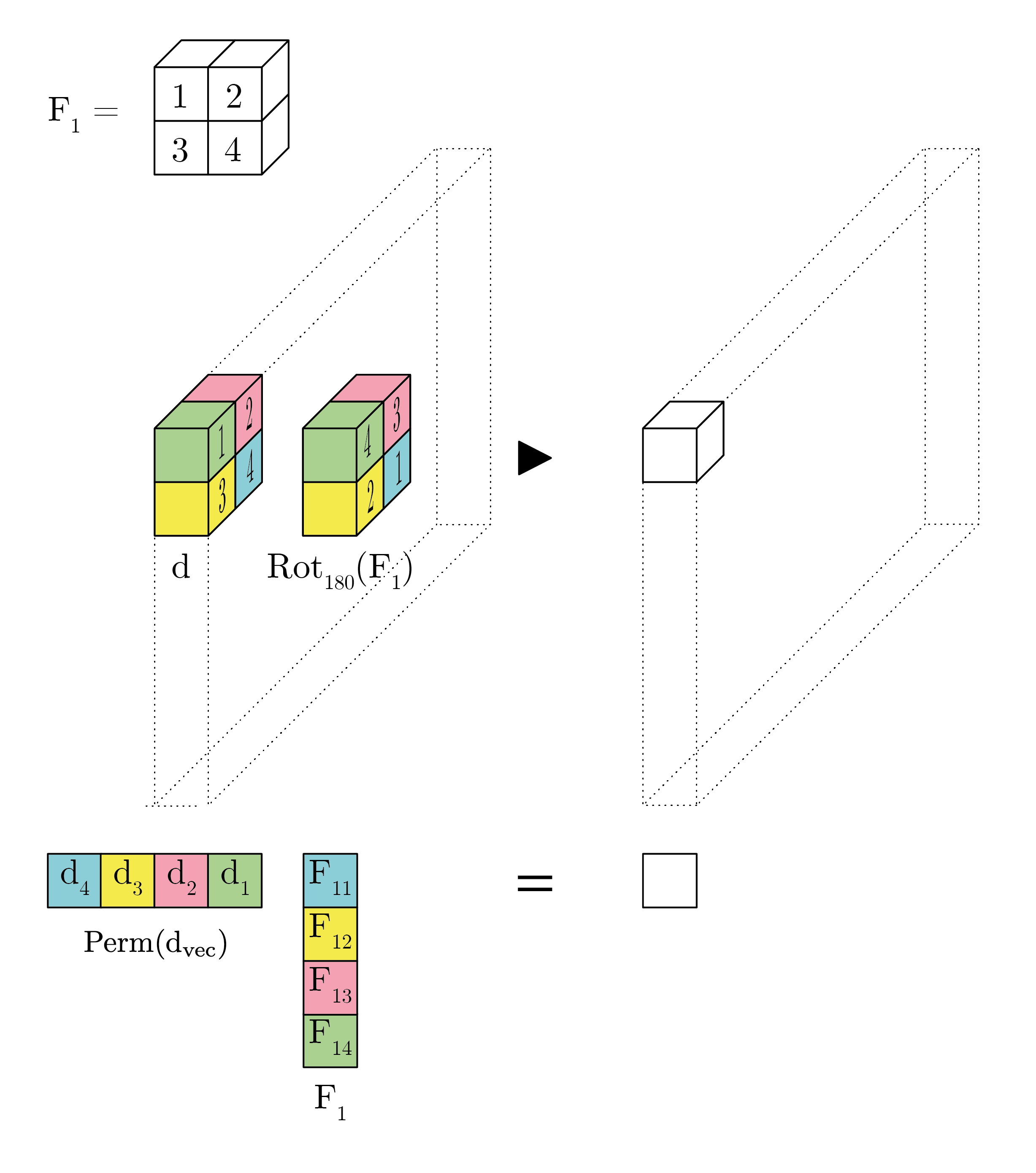}
     \caption{Back-propagation involves inner products with a layer's filters. In this figure, $d$ represents a patch of $\frac{\partial L}{\partial y}$, and $F$ is one of the layers convolutional filters.}
     \label{fig:dctps conv backprop}
 \end{figure}

This applies equally for the backward-pass, since backpropagation through convolutional layers involves convolutions as well. Let $h$ be the layer input, $y$ be the output of a layer with a single $2 \times 2$ filter $F$, and let $L$ be the loss. Calculating $\frac{\partial L}{\partial h}$ involves calculating $\frac{\partial L}{\partial y} * \text{Rot}_{180}(F)$. Again, each step in this convolution is equivalent to an inner product between the original filter, and a flattened, \textit{permuted} patch of $\frac{\partial L}{\partial y}$, see Figure \ref{fig:dctps conv backprop}. This generalises for larger filters \cite{boue2018deep}.

\section{Parameter Allocation by SynFlow}
\begin{figure}[h!]
     \centering
     \includegraphics[width=0.45\textwidth]{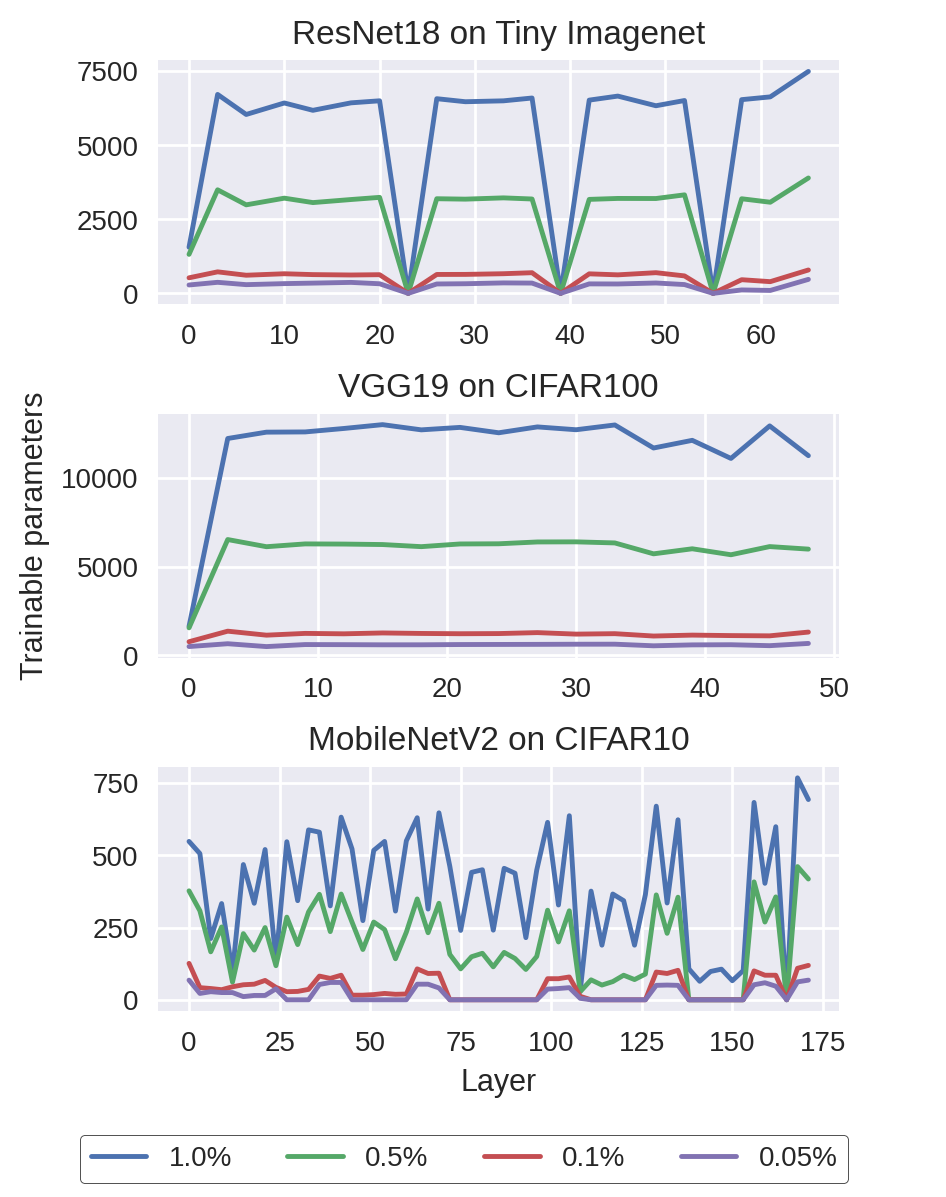}
     \caption{Total number of trainable parameters per layer, after pruning at initialization with SynFlow.}
     \label{fig:synflow nonzeros}
 \end{figure}

It was noted in Section 6 that when applied to ResNet50 for CIFAR10 at modest sparsities, SynFlow pruned fully those residual connections which were prunable (those implemented as trainable,  $1 \times 1$ convolutions), and in the remaining layers it appeared to approximate the EPL heuristic. Figure \ref{fig:synflow nonzeros} shows that this observation also applies to other architectures with different numbers of output classes. It is particularly striking the extent to which SynFlow applied to VGG19 leaves unpruned an approximately equal number of parameters per layer. On ResNet18, the pruning is observed to have the same structure per layer as on ResNet50 - a roughly equal number of trainable parameters per layer, except for the prunable shortcut connections, which are pruned completely. This behaviour is also present to some extent on MobileNetV2, though with larger oscillations around a central value, and a breakdown of this behaviour at lower densities, at which point multiple layers begin to be pruned in their entirety. 

\section{Cases where SynFlow and FORCE cannot be applied}
\subsection{Extreme Sparsities (FORCE)}

In some cases FORCE is unable to successfully prune past a certain sparsity. In particular, at some point in the pruning schedule, FORCE begins to assign all parameters a saliency score of 0, thus providing no basis for pruning, with the consequence being that the algorithm simply returns a dense network. This happens, for example, at the most extreme sparsities in VGG19, ResNet18, and MobileNetV2. In these cases, the test accuracy for FORCE is reported as equal to that of random guessing, since the algorithm cannot provide a trainable network at the given sparsity, but this is denoted with a dashed lined to indicate that no network of the specified sparsity was actually tested. 

We conjecture that this phenomenon is a result of \textit{throughput} collapse\footnote{This is equivalent to layer collapse when there is only a single feedforward connection at each layer.} - that is, once the algorithm fully prunes \textit{all branches} of communication at some point in the network, though we did not investigate this further. We note that in investigating these collapses, we also tried doubling the number of pruning steps from 60 to 120, but this did not avoid the problem.

\subsection{Fixup ResNet}
Neither FORCE nor SynFlow can be applied without modification to FixupResNet110. As above, this failure is due to the fact that both algorithms assign a saliency score of 0 to all parameters and consequently have no basis on which to prune. 

In particular, at initialization, the only non-zero gradients of both the training loss $\mathcal{L}$ and SynFlow's objective function $\mathcal{R}$, are in the network's final layer, where the weights themselves are initialized as 0. The saliency scores in both FORCE and SynFlow are obtained via the elementwise multiplication of the parameter matrices with their gradients, and thus are 0 in all cases.

\section{Experimental Details}

\subsection{Code and Implementation}
We implemented Force\footnote{\href{https://github.com/naver/force}{https://github.com/naver/force}} and SynFlow \footnote{\href{https://github.com/ganguli-lab/Synaptic-Flow}{https://github.com/ganguli-lab/Synaptic-Flow}} using the code published by the respective authors, adapted to include any additional architectures used in our experiments. For RigL, we adapted its PyTorch Implementation\footnote{\href{https://pypi.org/project/rigl-torch/}{https://pypi.org/project/rigl-torch/}}.

The code used to run experiments with DCTpS networks is available at \href{https://github.com/IlanPrice/DCTpS}{github.com/IlanPrice/DCTpS}.

\subsection{Model Architectures}
Standard implementations of network architectures used here are taken from the following sources:
\vspace{-3mm}
\begin{itemize}
    \item ResNet50, ResNet18 and VGG19, as implemented in \cite{jorge2021progressive}. See \href{https://github.com/naver/force/blob/master/experiments/models.py}{the FORCE Github Repo}
    \vspace{-3mm}
    \item MobileNetV2 from the authors' published code \href{https://github.com/kuangliu/pytorch-cifar/blob/master/models/mobilenetv2.py}{here}.
    \vspace{-3mm}   
    \item FixupResNet110 from the authors' published code \href{https://github.com/hongyi-zhang/Fixup}{here}.
\end{itemize}

\subsection{Parameter Breakdown by Architecture}

See Table \ref{tab:params} for a breakdown of the prunable/non-prunable parameter totals in each of the architectures used in our experiments.

\begin{table}[h!]
\caption{Division of total parameters between weights (pruned) and bias and/or batchnorm (BN) parameters (not pruned) in the architectures used in our experiments, with 10 output classes.}
{\renewcommand{\arraystretch}{1.5} 
\vspace{3mm}
\begin{tabular}{lrrr}
\toprule
               & Weights            & Bias \& BN & Total    \\ \hline
ResNet50       & 23467712 & 53130     & 23520842 \\
VGG19          & 20024000 & 11018    & 20035018 \\
MobileNetV2    & 2261824  & 35098     & 2296922  \\
FixupResNet110 & 1719856  & 282      & 1720138  \\
ResNet18       & 11164352  & 9610      & 11173962\\

\bottomrule
\end{tabular}
}
\label{tab:params}
\end{table}

\subsection{Pruning hyperparameters}
 See Table \ref{tab:prune_hyperparams} for the hyperparameters used when applying FORCE and SynFlow. RigL is tested with $T_{end}$ set as 75\% of total iterations, the RigL $\alpha$ parameter set to 0.3, with all convolutional and linear layers sparsified, according to the ERK distribution. When RigL is combined with DCTpS networks, EPL is used instead of ERK.
 
\begin{table}[h!]
    \centering
    \caption{Pruning hyperparameters for FORCE and SynFlow. C10, C100, and TI stand for CIFAR10, CIFAR100, and Tiny Imagenet respectively.}
    \label{tab:prune_hyperparams}
    {\renewcommand{\arraystretch}{1.5} 
\vspace{3mm}
\begin{tabular}{r r r}
    \toprule
     & FORCE & SynFlow \\
    \hline 
    Prune Steps & 60 & 100 \\
    \hline
    \# Batches    & 1 (C10),  10 (C100), 20 (TI) & N/A  \\
    \hline
    Schedule  &  exp  & exp \\
    \bottomrule
    \end{tabular}
    }
\end{table}

\subsection{Training Details}

See Table \ref{tab:train_hyperparams} for the training hyperparameters used in our experiments in Section 5 and Appendix \ref{app: experiments}. On CIFAR10 and CIFAR100 \cite{krizhevsky2009learning}, 10\% of the training data is withheld as a validation set. The model with the maximum validation accuracy is selected as our final model, to be evaluated on the test set.  In the case of Tiny Imagenet \cite{wu2017tiny}, where there are no labels for the test set, the maximum validation accuracy obtained during training is reported. All experiments were run using Adam, except for those in Appendix \ref{app: sgd} in which SGD with momentum was used. 

\begin{table}[h!]
    \centering
        \caption{Training hyperparameters used for experiments in Section 5 and Appendix \ref{app: experiments}. Note that for training DCTpS networks with SGD, a base learning rate of 0.03  was used instead of 0.1. For experiments with Lenet-5 (only performed with Adam on CIFAR10) batch size was 64 and total epochs was 160. }
    \label{tab:train_hyperparams}
    {\renewcommand{\arraystretch}{1.5} 
\vspace{3mm}
\begin{tabular}{lrr}
\toprule
     & Adam  & SGD \\
    \hline 
    Epochs     & 200 & 200 \\
    \hline
    Batch Size     & 128 & 128 \\
    \hline
    Learning Rate (LR)  &  0.001  & 0.1 \\
    \hline
    Momentum     & N/A & 0.9 \\
    \hline
    LR Decay Epochs & N/A & 120, 160 \\
    \hline
    LR Drop factor   & N/A & 0.1 \\
    \hline
    Weight Decay & $5 \times 10^{-4}$ & 5 $\times 10^{-4}$  \\
\bottomrule
    \end{tabular}
    }
\end{table}




\end{document}